\documentclass[letterpaper, 10 pt, conference, twoside]{ieeeconf} 
\usepackage{graphicx}
\usepackage{amsmath}
\usepackage{bbding}
\usepackage{pifont}
\usepackage{wasysym}
\usepackage{amssymb}
\usepackage{hyperref}
\usepackage{threeparttable} 
\usepackage{cite} 
\usepackage{booktabs}
\usepackage{xcolor}
\usepackage{multirow}
\usepackage{blindtext}

\usepackage{bm}
\usepackage{subcaption}
\usepackage[linesnumbered, ruled]{algorithm2e}

\SetKwInput{KwInput}{Input}
\SetKwInput{KwOutput}{Output}

\usepackage{cite}

\IEEEoverridecommandlockouts                              

\SetKwComment{Comment}{$\triangleright$\ }{}

\usepackage{afterpage}
\usepackage{mwe}
\title{\LARGE \bf 
CP-loss: Connectivity-preserving Loss for Road Curb Detection in Autonomous Driving with Aerial Images   
}

\author{Zhenhua Xu, \IEEEmembership{Student Member, IEEE}, Yuxiang Sun, \IEEEmembership{Member, IEEE}, Lujia Wang, \IEEEmembership{Member, IEEE}, \\and Ming Liu, \IEEEmembership{Senior Member, IEEE} 
\thanks{This work was supported by Zhongshan Municipal Science and Technology Bureau Fund, under project ZSST21EG06, Collaborative Research Fund by Research Grants Council Hong Kong, under Project No. C4063-18G, and Department of Science and Technology of Guangdong Province Fund, under Project No. GDST20EG54, awarded to Prof. Ming Liu.
}
\thanks{Zhenhua Xu is with the Department of Computer Science and Engineering, The Hong Kong University of Science and Technology, Clear Water Bay, Kowloon, Hong Kong (email: zxubg@connect.ust.hk).}
\thanks{Yuxiang Sun is with the Department of Mechanical Engineering, The Hong Kong Polytechnic University, Hung Hom, Kowloon, Hong Kong (email: yx.sun@polyu.edu.hk, sun.yuxiang@outlook.com).}
\thanks{Lujia Wang and Ming Liu are with the Department of Electronic and Computer Engineering, The Hong Kong University of Science and Technology, Clear Water Bay, Kowloon, Hong Kong (email: {eewanglj,eelium}@ust.hk).}
\thanks{\textit{Corresponding author:  Lujia Wang.}}}

\begin{document}


\maketitle
\begin{abstract}
Road curb detection is important for autonomous driving. It can be used to determine road boundaries to constrain vehicles on roads, so that potential accidents could be avoided. Most of the current methods detect road curbs online using vehicle-mounted sensors, such as cameras or 3-D Lidars. However, these methods usually suffer from severe occlusion issues. Especially in highly-dynamic traffic environments, most of the field of view is occupied by dynamic objects. To alleviate this issue, we detect road curbs offline using high-resolution aerial images in this paper. Moreover, the detected road curbs can be used to create high-definition (HD) maps for autonomous vehicles. Specifically, we first predict the pixel-wise segmentation map of road curbs, and then conduct a series of post-processing steps to extract the graph structure of road curbs. To tackle the disconnectivity issue in the segmentation maps, we propose an innovative connectivity-preserving loss (CP-loss) to improve the segmentation performance. The experimental results on a public dataset demonstrate the effectiveness of our proposed loss function. This paper is accompanied with a demonstration video and a supplementary document, which are available at \texttt{\url{https://sites.google.com/view/cp-loss}}. 



\end{abstract}
\section{Introduction}
Detection of road curbs is a fundamental task for autonomous driving. It can be used to determine the road boundaries, so that autonomous vehicles can be constrained on roads to avoid potential accidents, which is critical to autonomous driving safety. Different from regular road objects (e.g., vehicles and pedestrians), road curb is of line-shaped and usually thin and long. So they cannot be precisely detected by object detection methods \cite{he2017mask} that annotate objects with bounding boxes. In the past, vehicle-mounted sensors, such as Lidar \cite{wang2019point,zhang2018road} or camera \cite{panev2018road,oniga2008curb,siegemund2010curb}, are usually used to detect road curbs online by model-based methods. However, online road curb detection could be seriously affected by the occlusion issue.
For example, in highly-dynamic traffic environments, the camera filed of view could be severely occluded by dynamic road objects, so that there is insufficient curb information, and hence the detection performance could be degraded.
Thus, in this paper, we detect road curbs using bird-eye-view (BEV) images by semantic segmentation in an offline manner. The offline detection results could assist the autonomous vehicle as prior information, and they can also be used to build high-definition (HD) maps so that the large-scale deployment of autonomous vehicles could be accelerated.

To be best of our knowledge, there is currently no published work on offline road curb detection using BEV images by segmentation. But similar works can be found in the filed of road-network extraction \cite{bastani2018roadtracer,mattyus2017deeproadmapper} and lane detection \cite{homayounfar2018hierarchical,homayounfar2019dagmapper}. Some of these works obtain BEV images from the pre-built point-cloud map, while other works rely on aerial images captured by satellite or unmanned aerial vehicle cameras. Although images obtained from the pre-built point-cloud map are more precise, the point-cloud map is time-consuming and expensive to build and update. Thus, with high-resolution aerial images becoming more and more world-widely available, we propose to detect road curbs from aerial images in this paper.

Line-shaped objects like lanes and road-networks are mainly detected in two steps: (1) Predict the pixel-wise semantic segmentation map of the target objects and then (2) conduct a series of post-processing such as thresholding and skeletonization \cite{mattyus2017deeproadmapper,batra2019improved,neven2018towards} on the segmentation map and obtain the graph structure of the target objects. In this paper, we focus on the first step (i.e., semantic segmentation of road curbs). Our proposed segmentation method can be treated as a module and applied to various line-shaped object detection solutions that may have different post-processing algorithms and pipelines.
The correctness of the segmentation map is critical to ensure satisfactory performance, including pixel-level correctness and topological correctness.
Since semantic segmentation directly works on image pixels, the spatial information of images is usually ignored and there lacks constraints on connections between pixels. As a result, the segmentation results tend to suffer from disconnectivity issue, which is hard to be effectively handled by hard-code post-processing algorithms. Therefore, the semantic segmentation method should be enhanced from the perspective of connectivity. 


The cross-entropy loss ($L_{CE}$) is a widely used pixel-level loss function for semantic segmentation. Although it is suitable for common segmentation tasks, it does not have a satisfactory performance for line-shaped object segmentation due to its intrinsic local property. 
Focal loss ($L_{FL}$) \cite{lin2017focal} can relieve the disconnectivity issue to some extent by assigning harder samples with higher weights, but it still fails to explicitly penalize disconnectivity. Different from the above two pixel-level loss functions, dice loss ($L_{Dice}$) \cite{milletari2016vnet} can capture the structural information from a global scale and directly optimize on the evaluation metric. However, 
$L_{Dice}$ cannot locate pixels affected by disconnectivity, either. Inspired by \cite{deng2018learning}, we combine $L_{Dice}$ and $L_{CE}$ for both pixel-level and image-level supervision.

In this paper, we propose a novel connectivity-preserving loss function (CP-loss) to relieve the aforementioned disconnectivity issue.
Our CP-loss can compare skeletons of the predicted segmentation result and the ground-truth label, then enlarge the training weights of pixels belonging to areas where the disconnection occurs. In this way, the semantic segmentation result will have better topological correctness and the disconnectivity issue of the final road curb graph structure can be greatly alleviated.
Our contributions are summarized here:
\begin{itemize}
    \item We propose an innovative loss function CP-loss, which enables the segmentation network to focus on regions of disconnectivity. This loss can be directly applied in existing line-shaped object detection solutions.
    \item We design a metric skeleton-connectivity measure (SCM) to evaluate the connectivity of the obtained road curbs.
    \item We perform extensive comparative studies to show that the proposed CP-loss outperforms all the other baseline loss functions.
\end{itemize}

\section{Related Works}
\subsection{Line-shaped object detection}
In recent years, line-shaped object detection has drawn great attention. The current works have multiple different goals, including road-network extraction \cite{bastani2018roadtracer,mattyus2017deeproadmapper,batra2019improved,he2020sat2graph}, lane detection \cite{homayounfar2018hierarchical,homayounfar2019dagmapper,neven2018towards,pan2017spatial,hou2019learning}, automatic annotation \cite{acuna2018efficient,castrejon2017annotating}, etc. Among them, some recent works detected line-shaped objects by iterative graph growing (i.e., grow the graph vertex by vertex) \cite{bastani2018roadtracer,xu2021topo,castrejon2017annotating,homayounfar2019dagmapper,xu2021icurb}, but they are limited to detecting objects with specific structures and suffer from accumulated errors. In addition, these methods are very slow due to the iterations. Another common category of method is semantic segmentation followed by post-processing. For example, Pan \textit{et al.} \cite{neven2018towards} proposed a new convolution approach to better extract spatial information of the input image, and they achieved a much higher evaluation score in the lane detection task. In \cite{mattyus2017deeproadmapper}, the authors first obtained the segmentation result containing many disconnections, then they designed an algorithm to generate candidate connections to bridge these disconnections, and trained another neural network to filter candidate connections. However, these methods mainly work on designing post-processing algorithms to refine the segmentation map while leaving improving the topological correctness of the segmentation map unexplored.
\subsection{Structure-aware semantic segmentation}
Past works from various research areas focus on designing segmentation methods that can capture the structural information of the input images. In \cite{hu2019topologypreserving,clough2019explicit}, the authors used Persistent Homology (PH) to explicitly evaluate the topology of the predictions, which obtained good results in simple scenarios. But PH only constraints the Betti number of the predictions and cannot capture the structural information of more complicated shapes, such as branches. Thus it cannot maintain good performance for complicated cases and is very time-consuming. Affinity learning \cite{oner2020promoting,affinity} is another category of method to enhance connectivity by finding the shortest paths between pixels. Oner \textit{et al.} \cite{oner2020promoting} proposed a novel loss function by calculating the shortest path between each pair of pixels of the background. If there exists disconnecivity of the predicted road-network, the loss function would rise. These works could explicitly define the connectivity so they tend to have impressive outcomes. However, they are usually time-consuming and only work for specific structures. Different from past works on pixel-level loss functions, Shit \textit{et al.} \cite{shit2020cldice} first made use of skeletons to emphasize connectivity. In this work, a differentiable soft-skeletonization algorithm and a soft-dice loss function based on obtained skeletons were proposed. But their method did not fully utilize the structural information of the skeletons considering that only a naive dice loss was implemented.

\section{The Proposed Method}
\begin{figure*}[t]
  \centering
    \includegraphics[width=\linewidth]{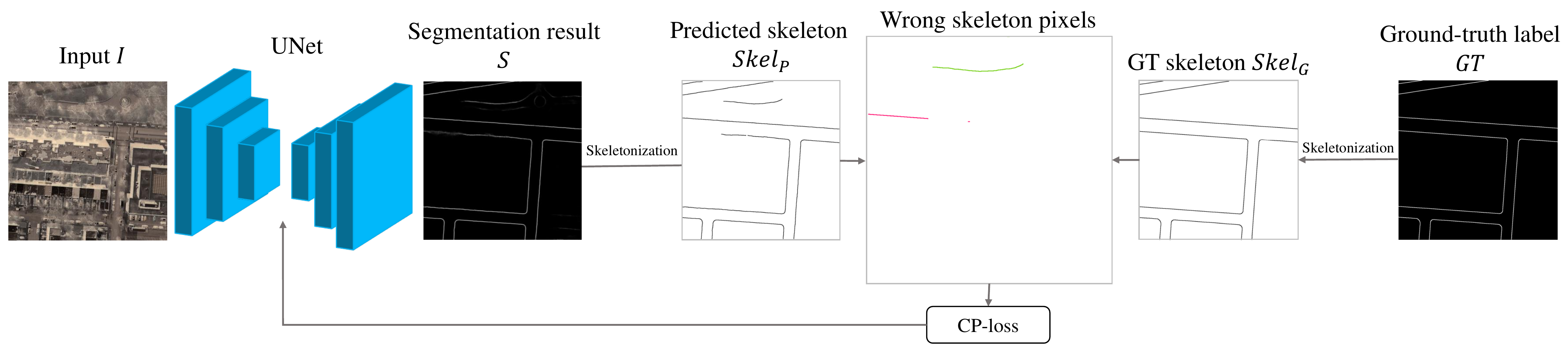}
  \caption{The overview of CP-loss. We first get the segmentation result $S$ from a segmentation network, then we skeletonize both $S$ as well as the ground-truth label $GT$. After that, the skeleton of $S$ as $Skel_{P}$ and the skeleton of $GT$ as $Skel_{G}$ are obtained. Please note that $GT$ is already of one-pixel-width, so $GT=Skel_{G}$. By comparing $Skel_{P}$ and $Skel_{G}$, we have the failed-retrieved ground-truth skeleton pixels denoted as $Skel_{fG}$ (red pixels), and the false-positive predicted skeleton pixels denoted as $Skel_{fP}$ (green pixels). $Skel_{fG}$ can be used to measure disconnectivity of $S$ and $Skel_{fP}$ is used to penalize ghost skeletons. The weight coefficients of our CP-loss are calculated based on $Skel_{fG}$ and $Skel_{fP}$. For better visualization, the lines in this figure are widened but they are actually of one-pixel-width. The figure is best viewed in color.}
  \label{diagram}
\end{figure*}
\subsection{Overview}
This work aims to detect road curbs in aerial images and relieve the disconnectivity issue of semantic segmentation. Our model takes as input an aerial image and outputs a skeleton of road curbs. In past semantic segmentation works, connectivity is not specially considered and emphasized, so the disconnectivity cannot be avoided and is hard to fix by hard-code post-processing. Cross-entropy loss ($L_{CE})$ is a pixel-level loss function and dice loss ($L_{Dice}$) is an image-level loss function. We combine them together to supervise the training, so that the network can capture both pixel-level and image-level information. However, both $L_{CE}$ and $L_{Dice}$ cannot explicitly define disconnectivity and effectively penalize it. Thus we need to find the regions where disconnectivity happens and make the network focus on them. To locate the disconnectivity, we extract skeletons of both the predicted segmentation map $S$ and the ground-truth label $GT$ as $Skel_P$ and $Skel_G$, respectively. Then the difference between two skeletons are measured based on Euclidean distance. After comparing two skeletons, we will penalize $Skel_{fG}$ (failed-retrieved ground-truth skeleton) and $Skel_{fP}$ (false-positive predicted skeleton). Both $L_{CE}$ and $L_{Dice}$ will be assigned with weights calculated from $Skel_{fG}$ and $Skel_{fP}$ to emphasize disconnectivity. Then $L_{CE}$ and $L_{Dice}$ are summed up as CP-loss, which is expressed in the following equation:
\begin{equation}
\begin{aligned}
L_{CP} =& L_{CE} + L_{Dice} \\
        =&\sum_i [-u_ig_ilog(p_i)-v_i(1-g_i)log(1-x_i)]\\
          &+(1-2\frac{\sum_i \beta_ip_ig_i}{\sum_i (\beta_ip_i)^2+\sum_ig_i^2}),
\end{aligned}
\end{equation}
where $g_i$ indicates the ground-truth value of a pixel $x_i$ and $p_i$ is the prediction of $x_i$. $u_i, v_i$ and $\beta_i$ are Euclidean distance-based weight coefficients to emphasize the connectivity. Details about the above equation is given in \ref{cploss}. Note that the ground-truth label is already of one-pixel-width, so $GT=Skel_{G}$ and the skeletonization on $GT$ can be omitted. The schematic overview of our method is visualized in Fig. \ref{diagram}. 

\subsection{Semantic Segmentation network}
This paper concentrates on loss function design, so we directly use a current representative deep network for binary segmentation (i.e., UNet \cite{ronneberger2015unet}). Since our loss function CP-loss does not have any requirement on network structures, it can be directly plugged into any semantic segmentation network to enhance the connectivity. 

\subsection{Skeletonization}
Skeletonization is a morphological operation in computer vision. It takes as input a binary image and outputs a one-pixel-width binary skeleton which represents the shape of the input. The obtained skeleton maintains the general shape and topology of the input binary image, so it can be used to represent the structure of the input image. Since skeletonization is used to calculate the weight coefficients of CP-loss, it does not have to be differentiable, thus we directly use the off-the-shelf library
\textit{skimage.morphology.skeletonize}\footnote{{\url{https://scikit-image.org/docs/dev/auto_examples/edges/plot_skeleton.html}}}. After we obtain the segmentation result $S$, we first threshold it by a threshold $\tau_{Bin}$ to generate a binary image. Then we extract the skeleton of $S$ as $Skel_{P}$ by skeletonization. We can get the skeleton of the ground-truth label $GT$ as $Skel_{G}$ in the same way. 

 \begin{figure*}[t]
 \centering
    \begin{subfigure}[t]{0.1375\textwidth}
        \begin{subfigure}[t]{\textwidth}
            \includegraphics[width=\textwidth]{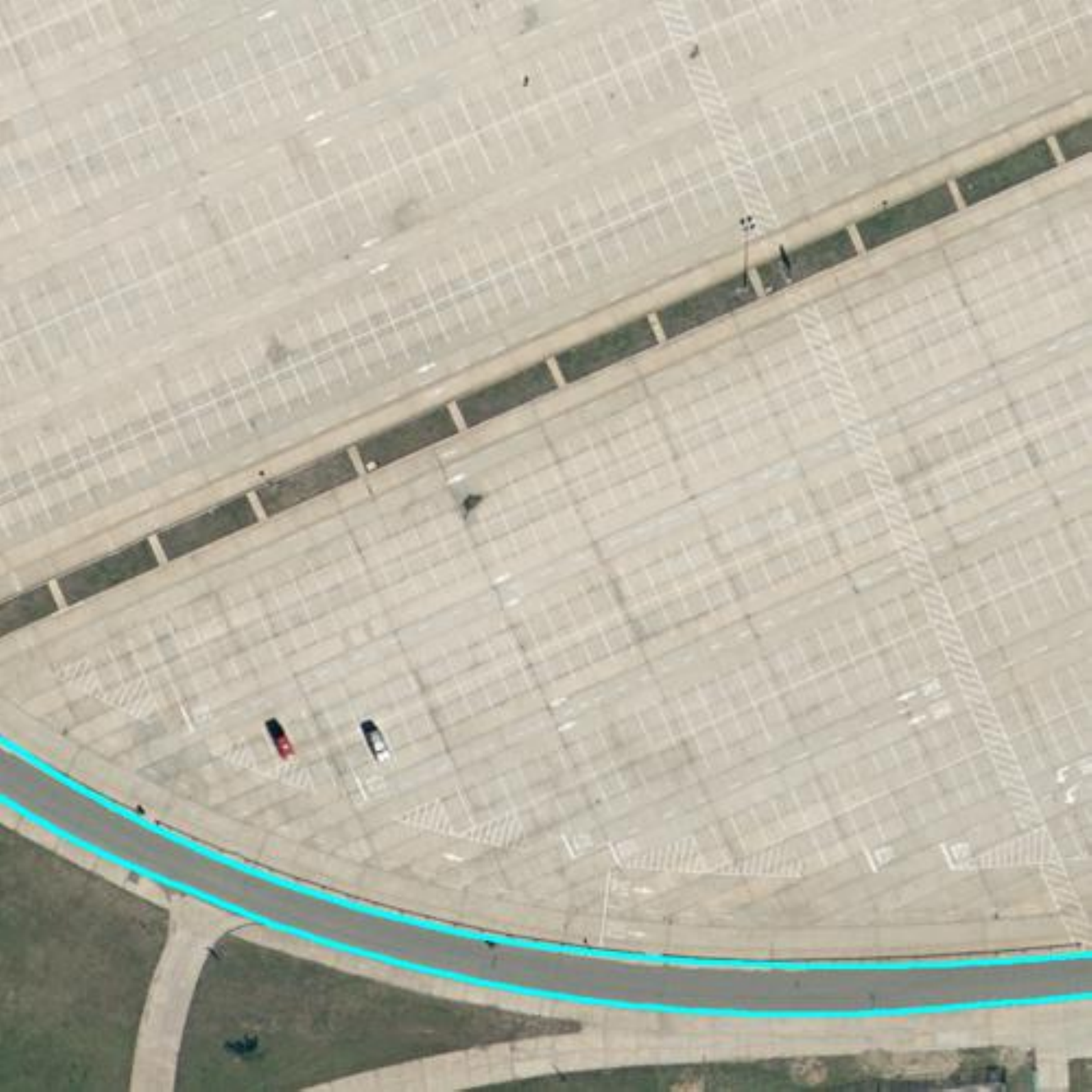}
        \end{subfigure}\vspace{.6ex}
        \begin{subfigure}[t]{\textwidth}
            \includegraphics[width=\textwidth]{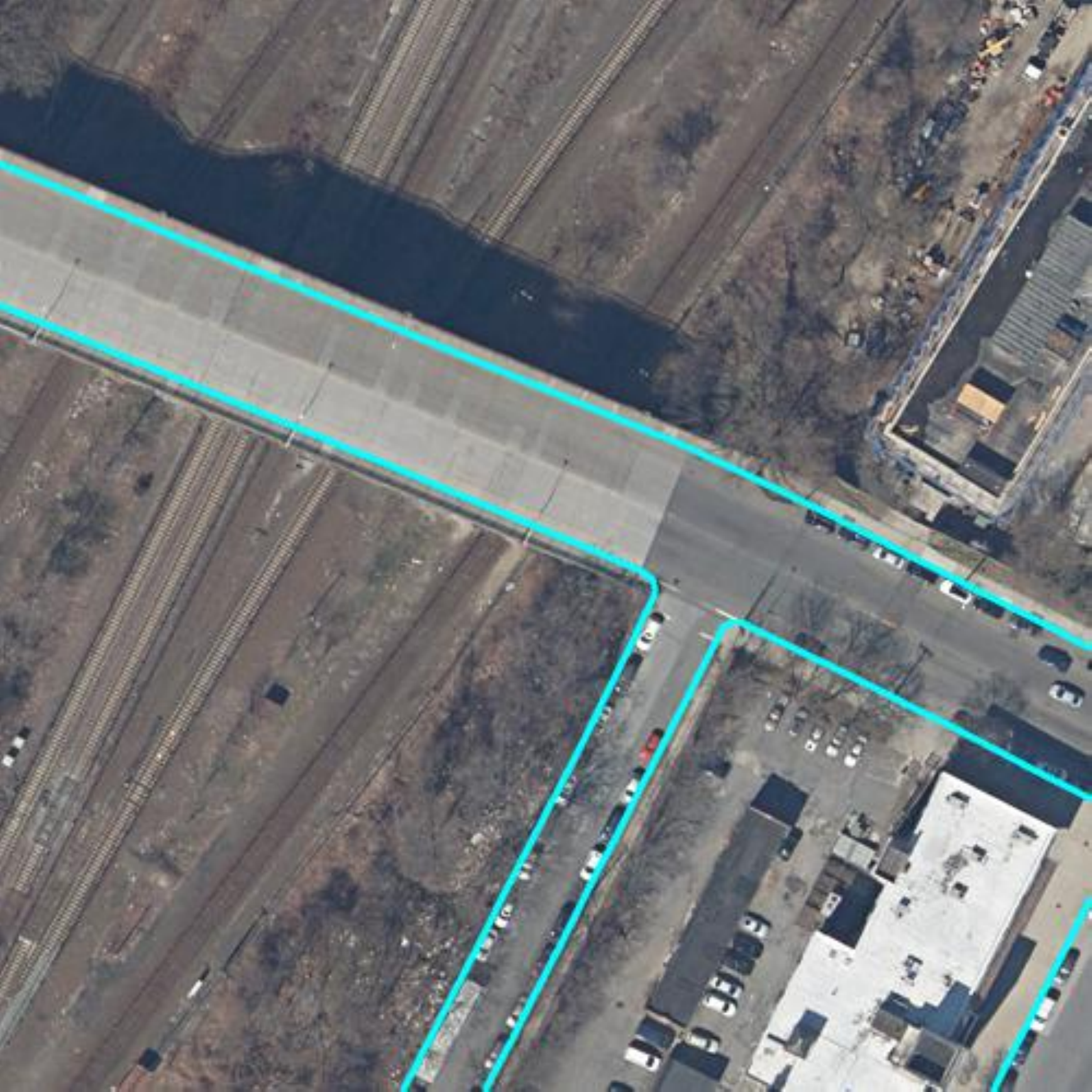}
        \end{subfigure}\vspace{.6ex}
        \begin{subfigure}[t]{\textwidth}
            \includegraphics[width=\textwidth]{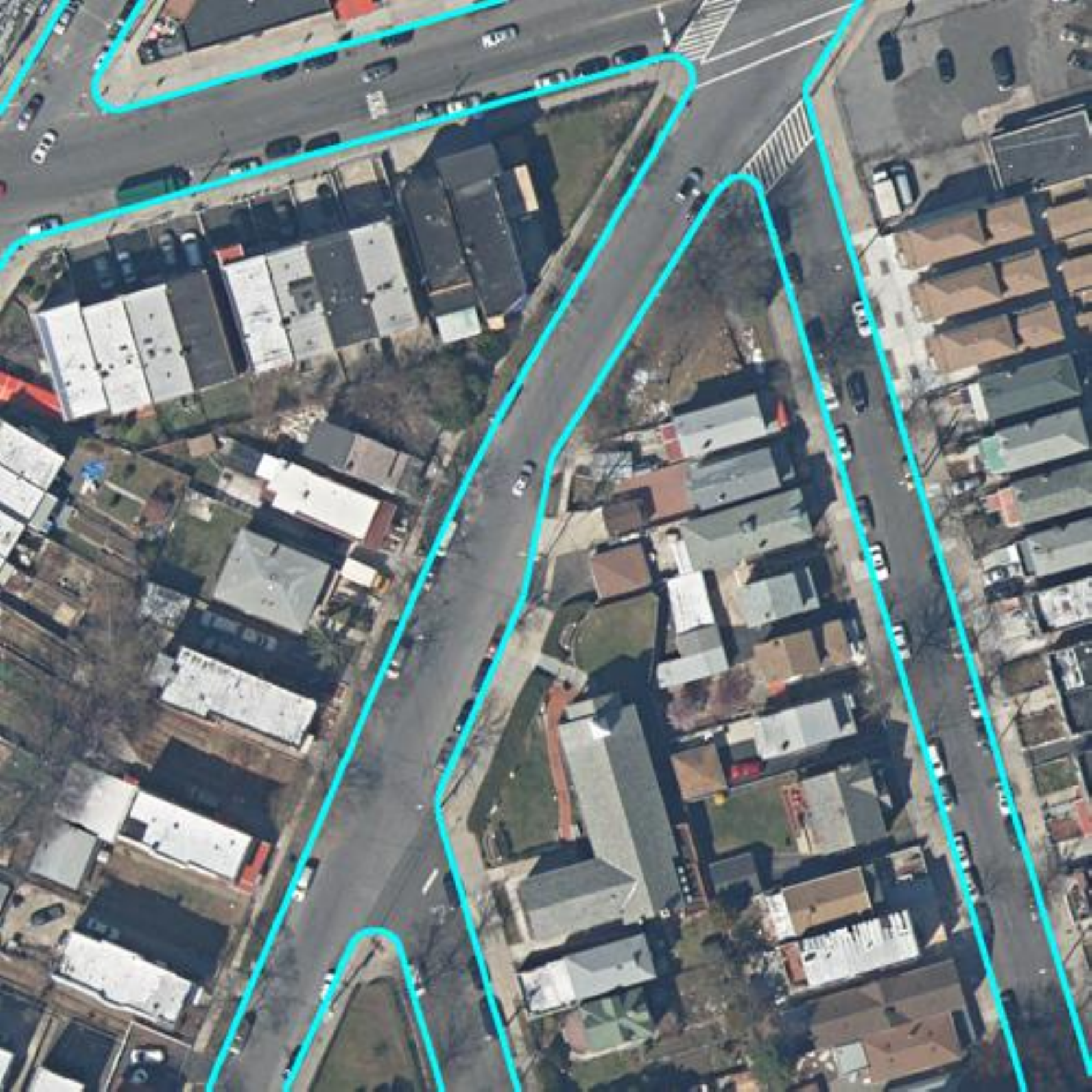}
        \end{subfigure}
        \caption{\footnotesize{Ground-truth}}
        \label{fig_qualitative_1st}
    \end{subfigure}
    \hfill
    \begin{subfigure}[t]{0.1375\textwidth}
        \begin{subfigure}[t]{\textwidth}
            \includegraphics[width=\textwidth]{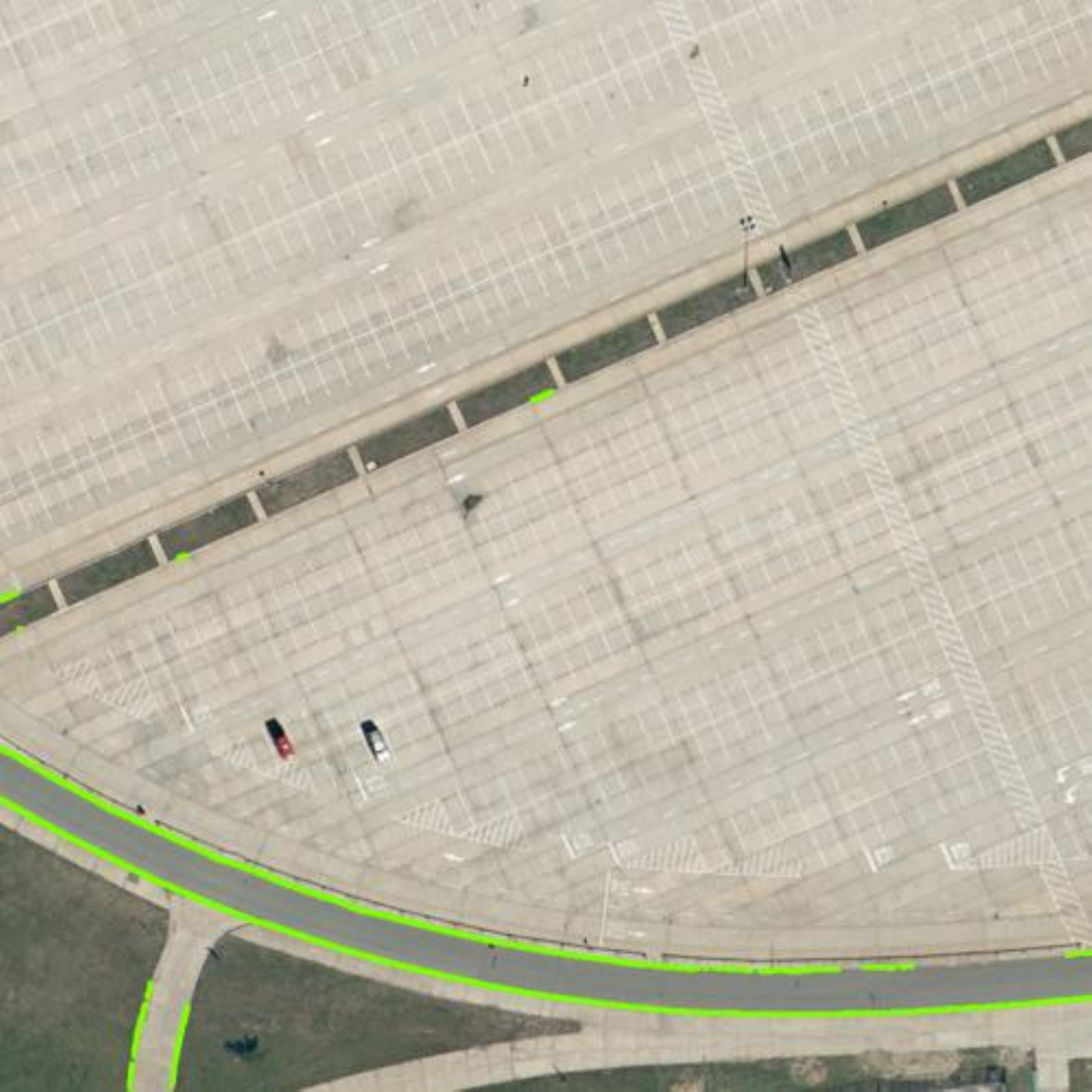}
        \end{subfigure}\vspace{.6ex}
        \begin{subfigure}[t]{\textwidth}
            \includegraphics[width=\textwidth]{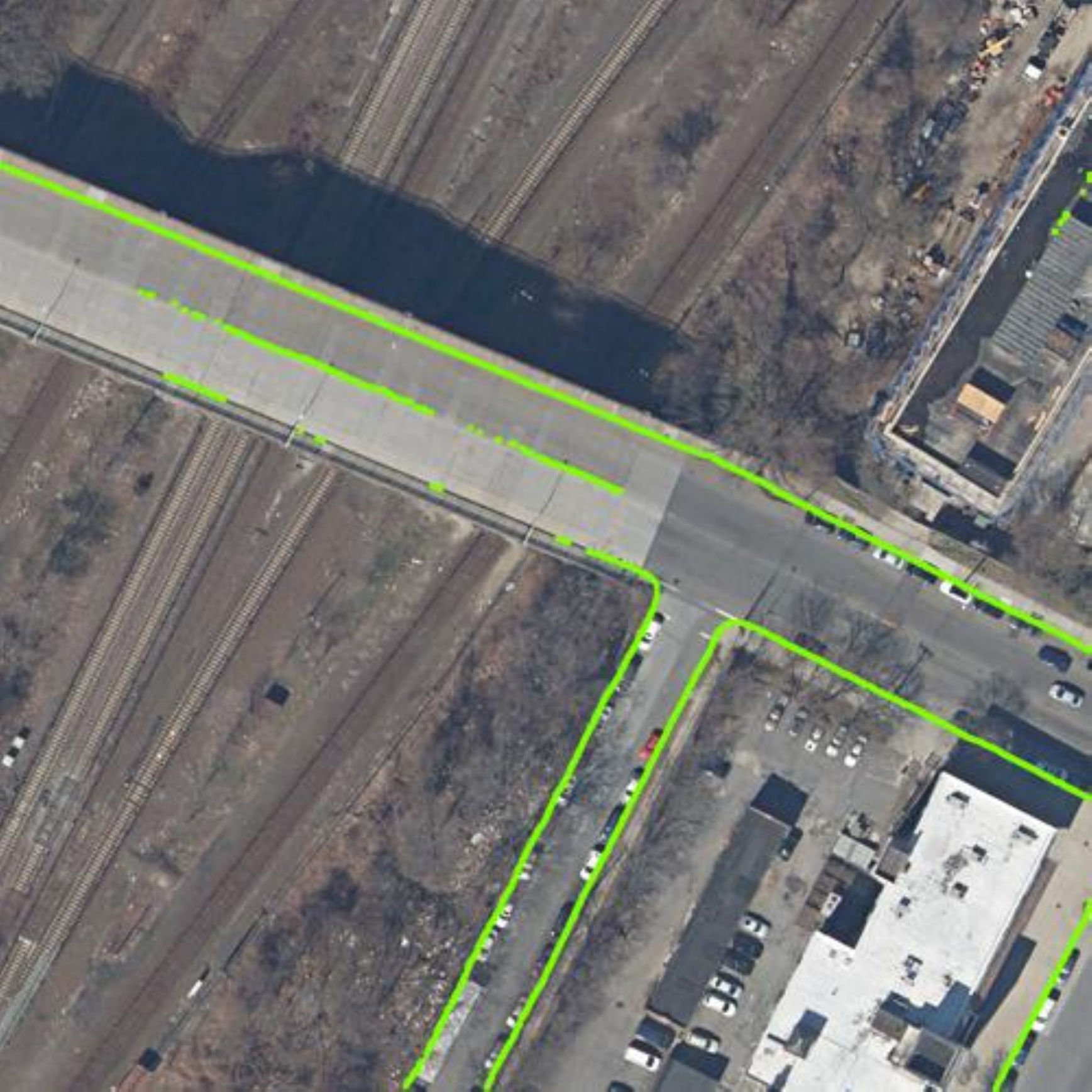}
        \end{subfigure}\vspace{.6ex}
        \begin{subfigure}[t]{\textwidth}
            \includegraphics[width=\textwidth]{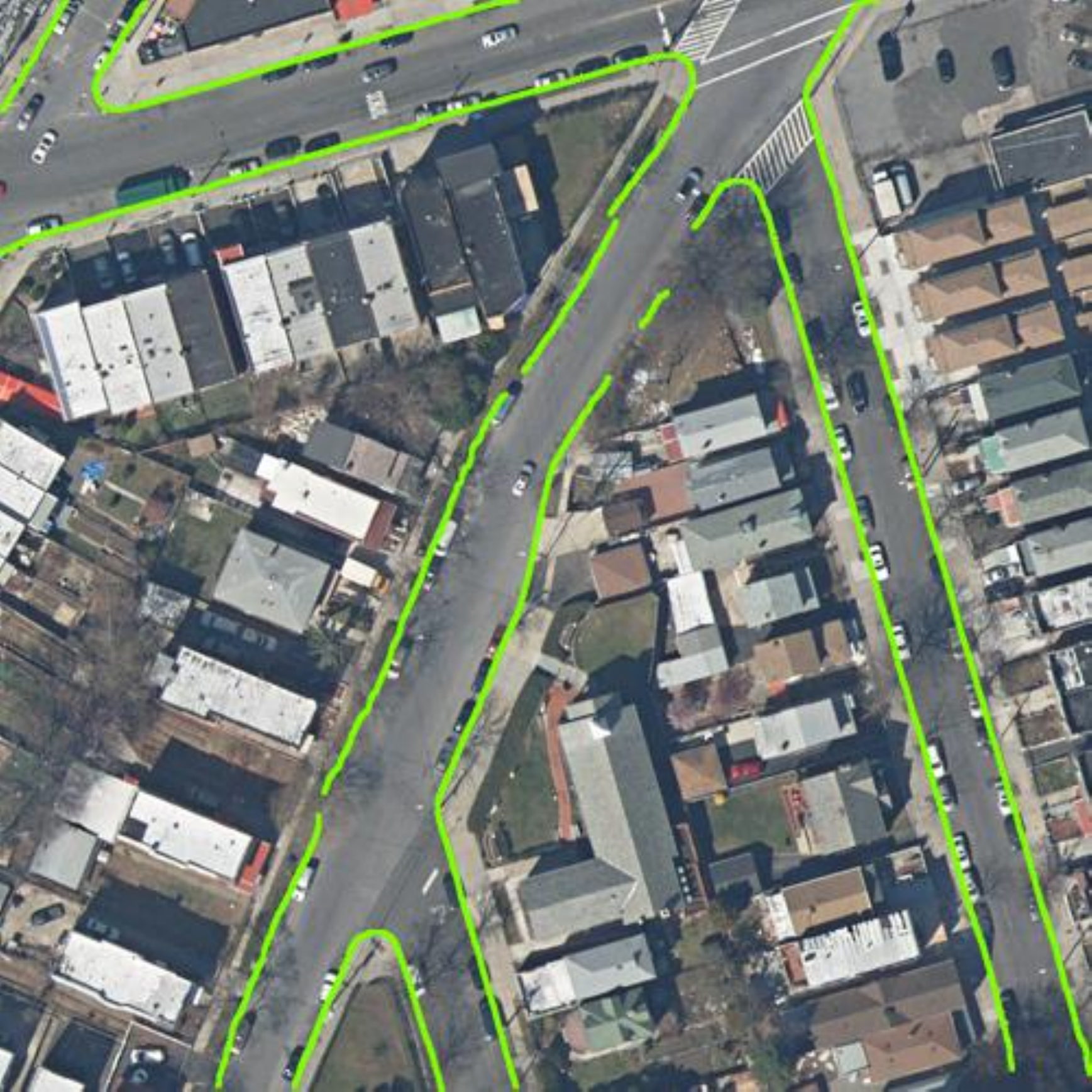}
        \end{subfigure}
        \caption{\footnotesize{BCE}}
    \end{subfigure}
    \hfill
    \begin{subfigure}[t]{0.1375\textwidth}
        \begin{subfigure}[t]{\textwidth}
            \includegraphics[width=\textwidth]{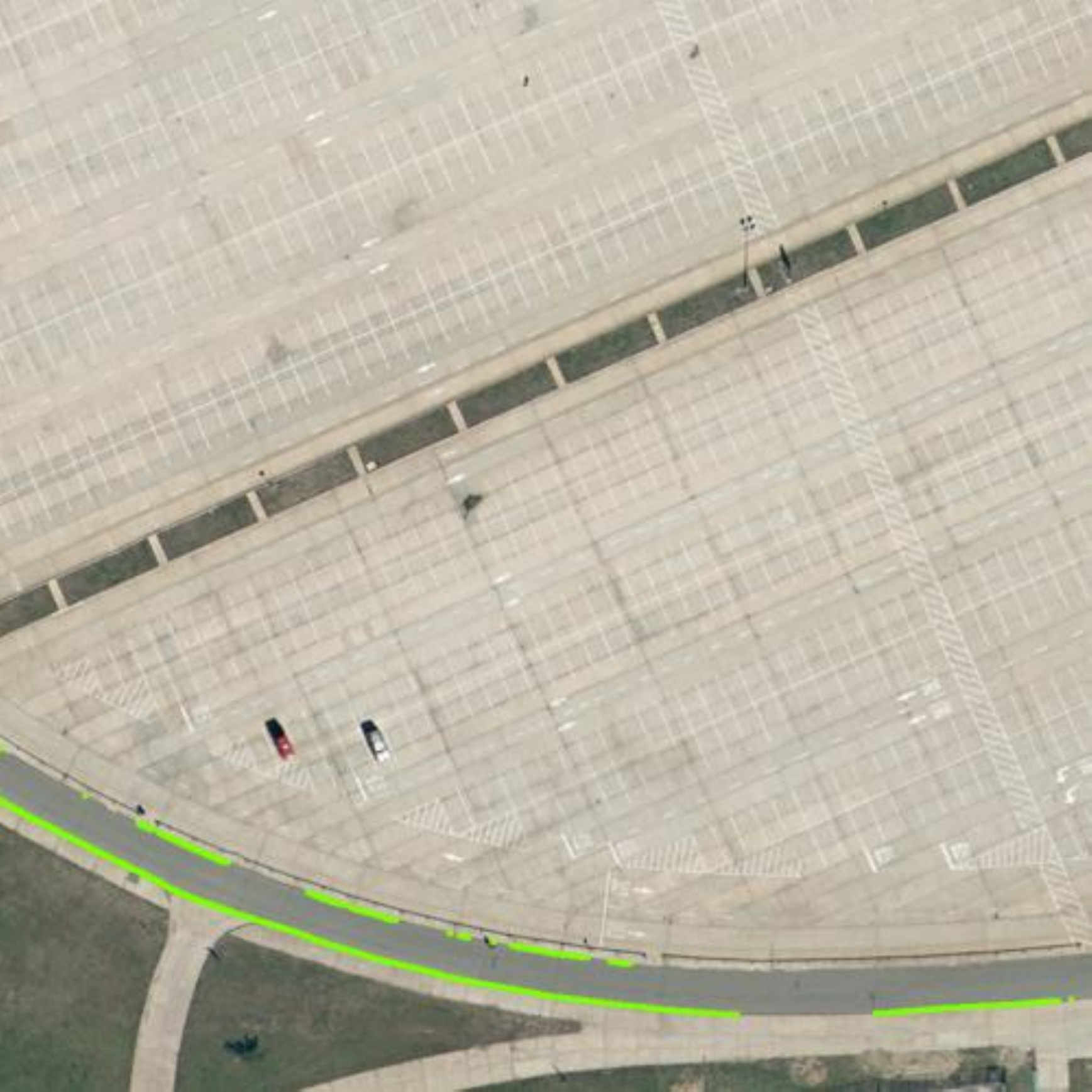}
        \end{subfigure}\vspace{.6ex}
        \begin{subfigure}[t]{\textwidth}
            \includegraphics[width=\textwidth]{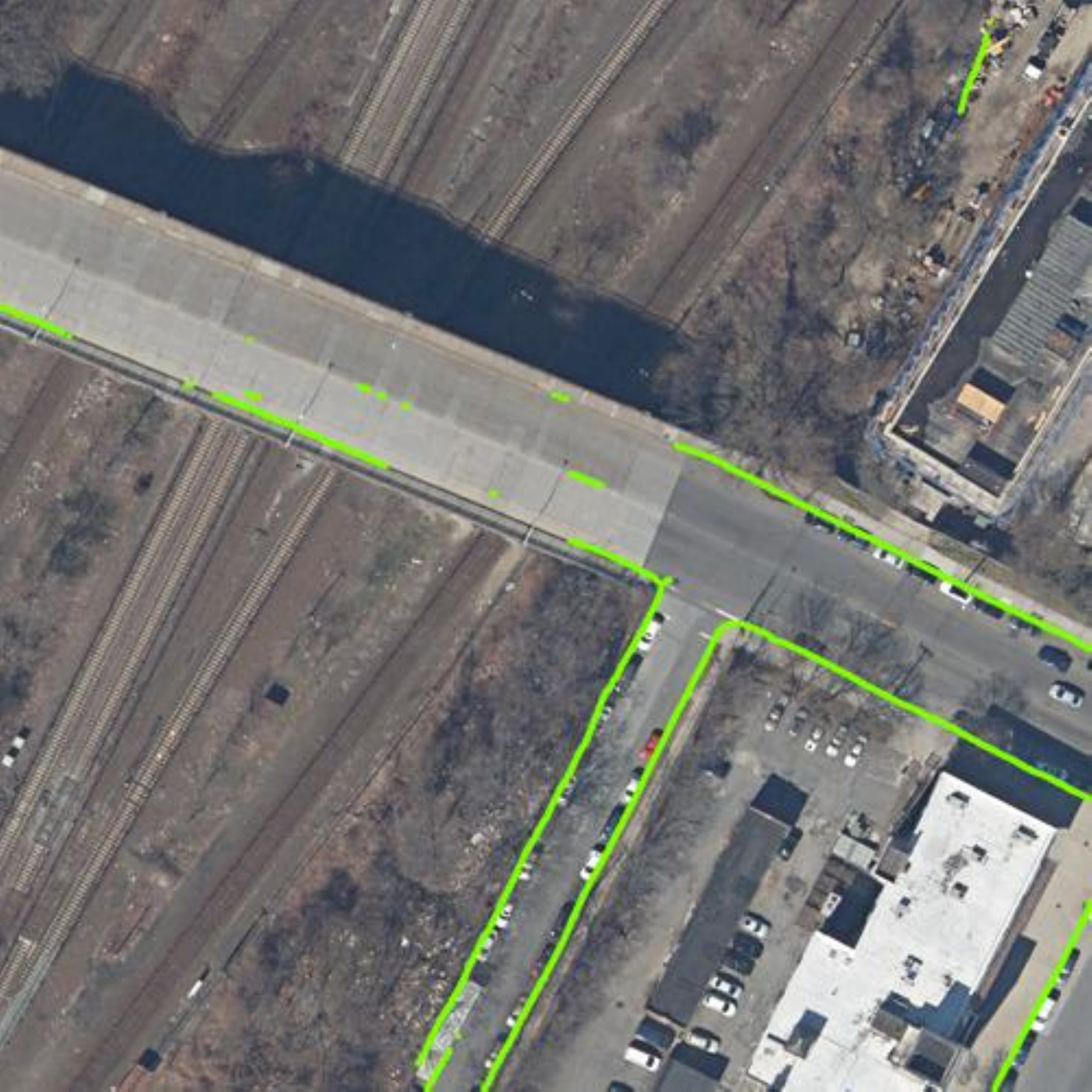}
        \end{subfigure}\vspace{.6ex}
        \begin{subfigure}[t]{\textwidth}
            \includegraphics[width=\textwidth]{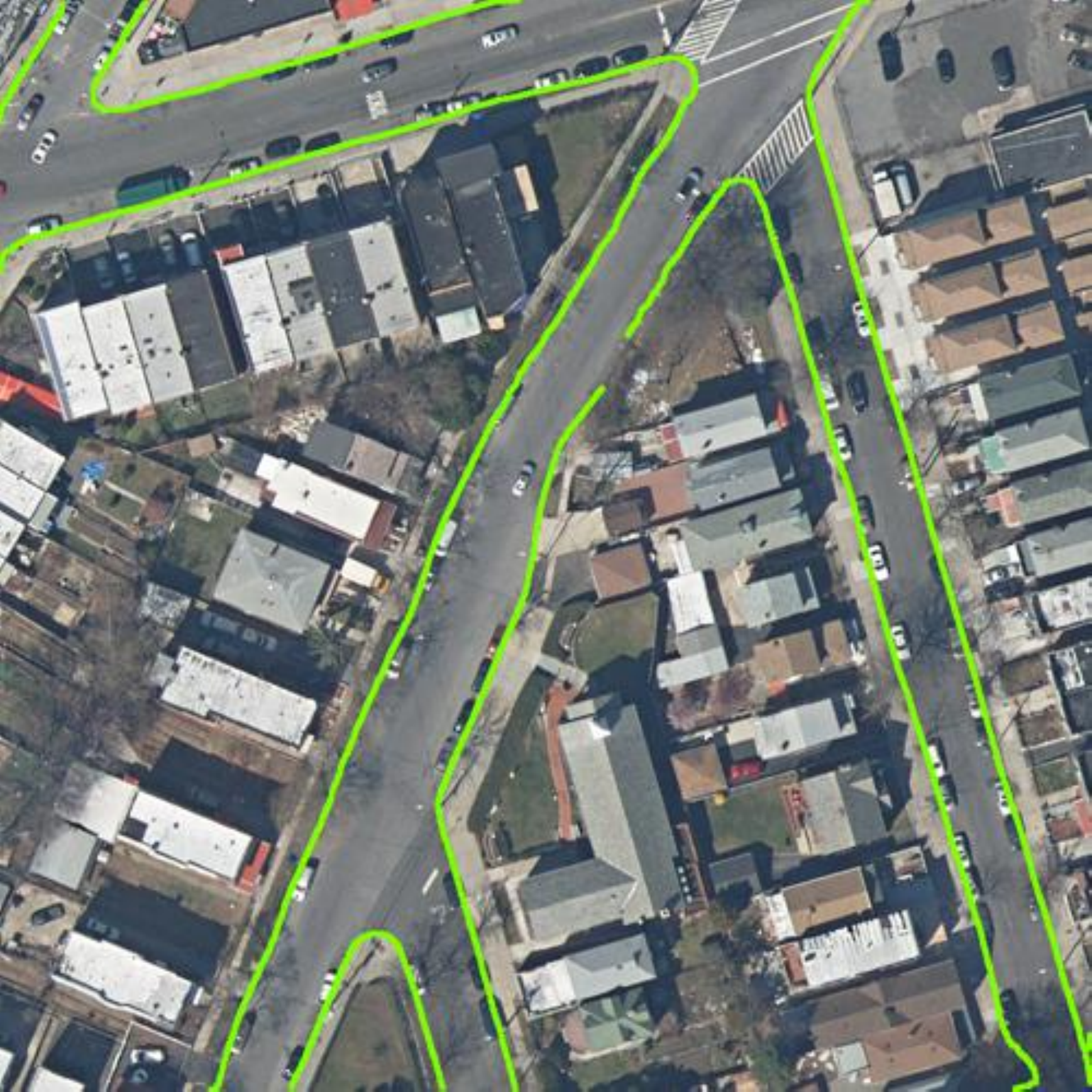}
        \end{subfigure}
        \caption{\footnotesize{Focal loss \cite{lin2017focal}}}
        \label{fig3_3}
    \end{subfigure}
    \hfill
    \begin{subfigure}[t]{0.1375\textwidth}
        \begin{subfigure}[t]{\textwidth}
            \includegraphics[width=\textwidth]{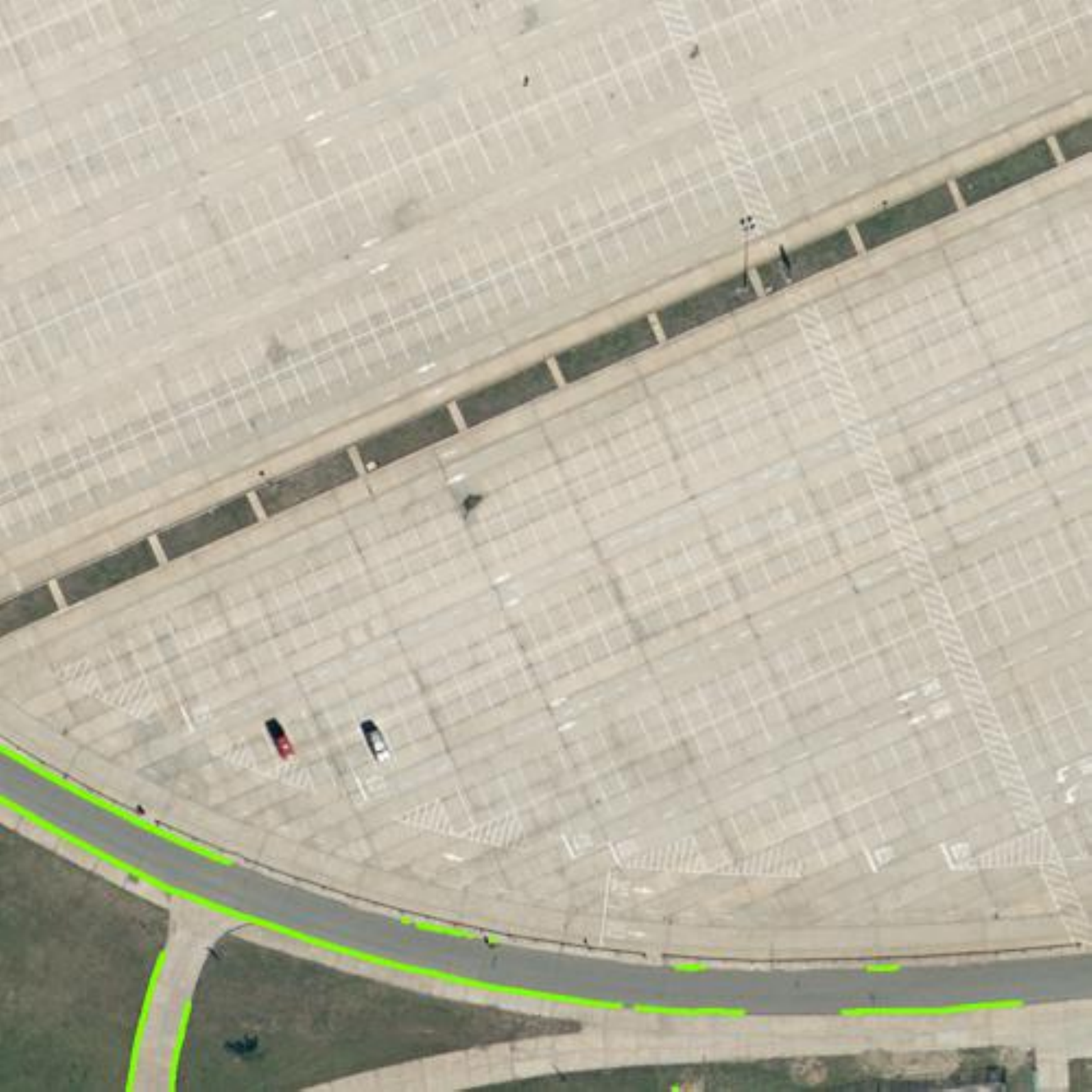}
        \end{subfigure}\vspace{.6ex}
        \begin{subfigure}[t]{\textwidth}
            \includegraphics[width=\textwidth]{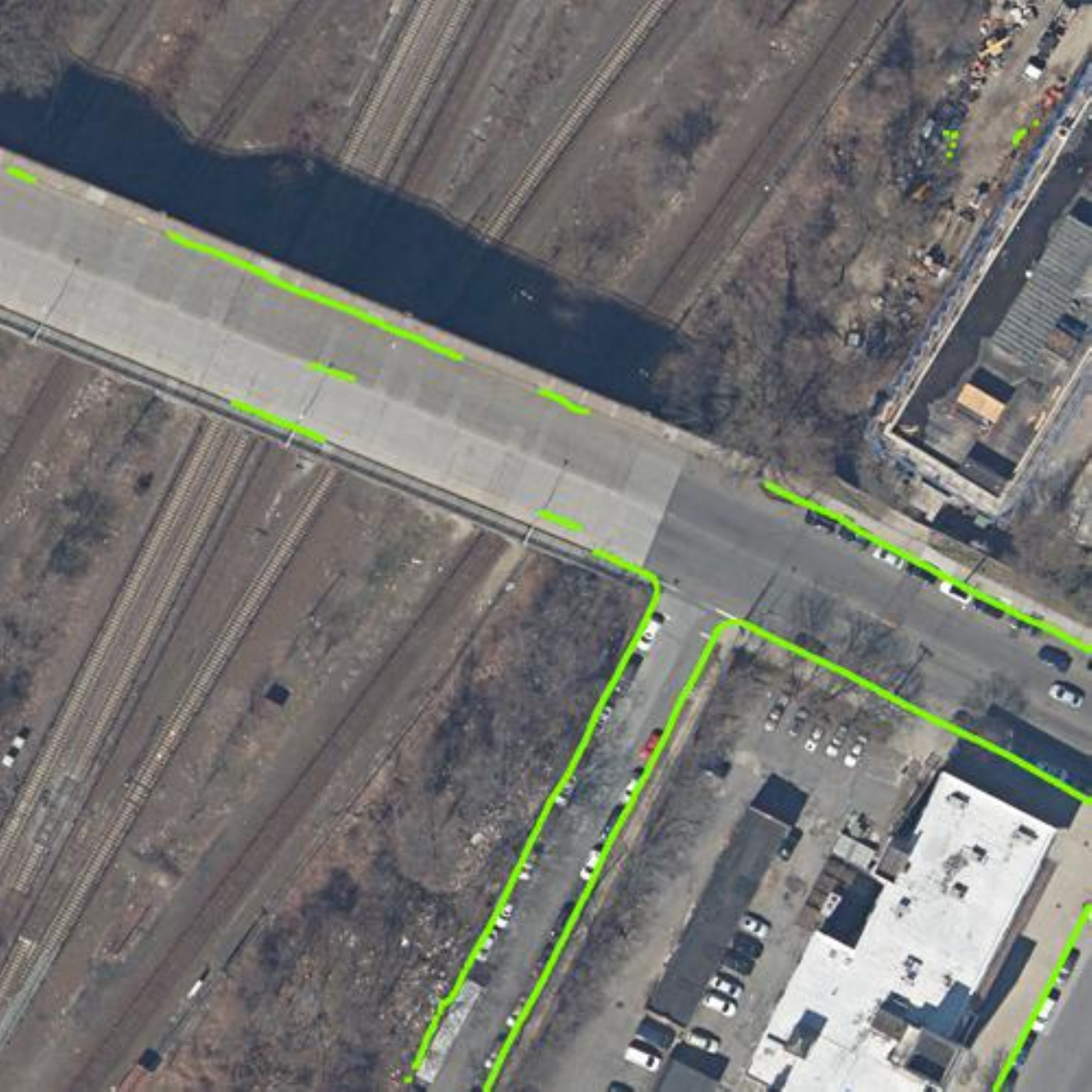}
        \end{subfigure}\vspace{.6ex}
        \begin{subfigure}[t]{\textwidth}
            \includegraphics[width=\textwidth]{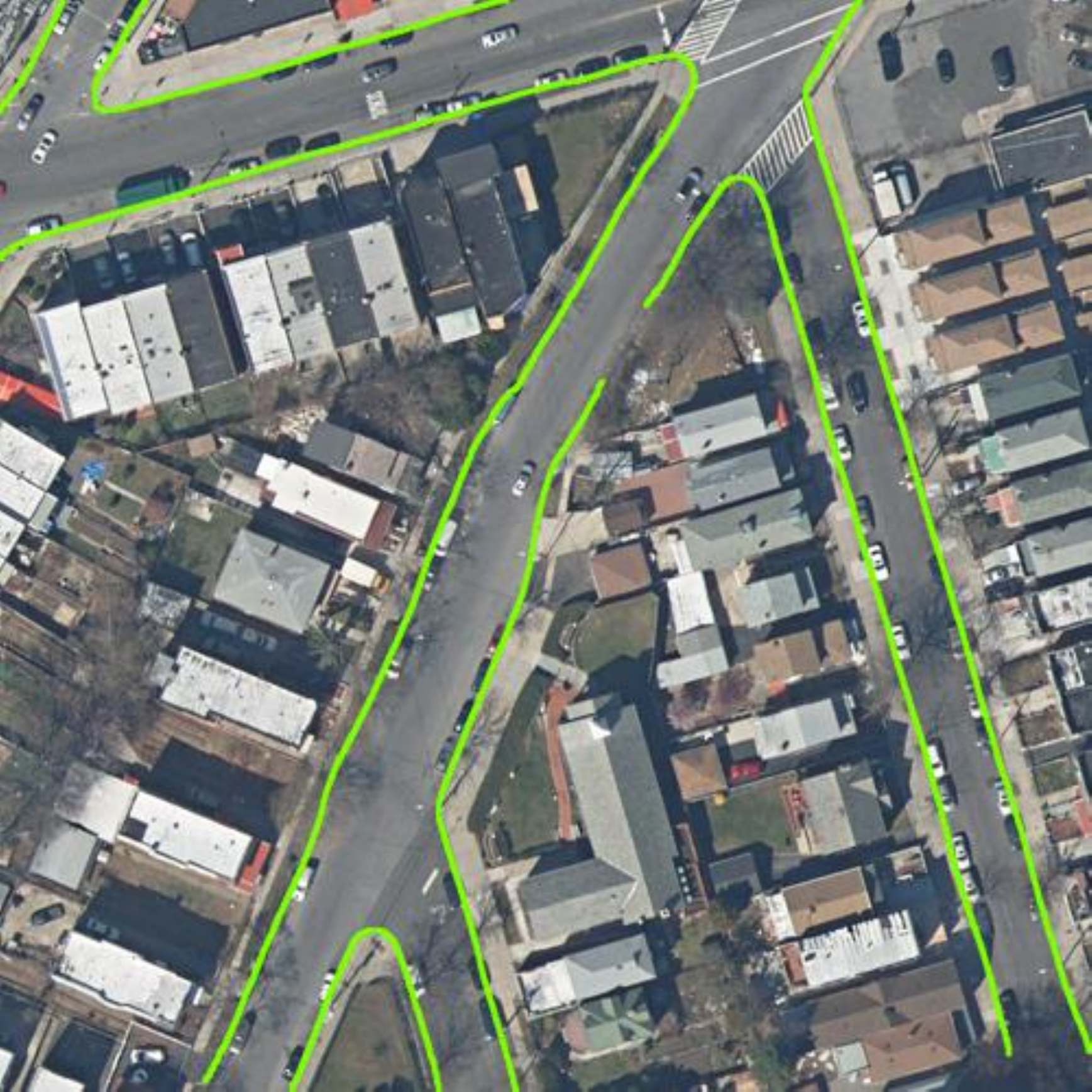}
        \end{subfigure}
        \caption{\footnotesize{Distance CE \cite{caliva2019distance}}}
    \end{subfigure}
    \hfill
    \begin{subfigure}[t]{0.1375\textwidth}
        \begin{subfigure}[t]{\textwidth}
            \includegraphics[width=\textwidth]{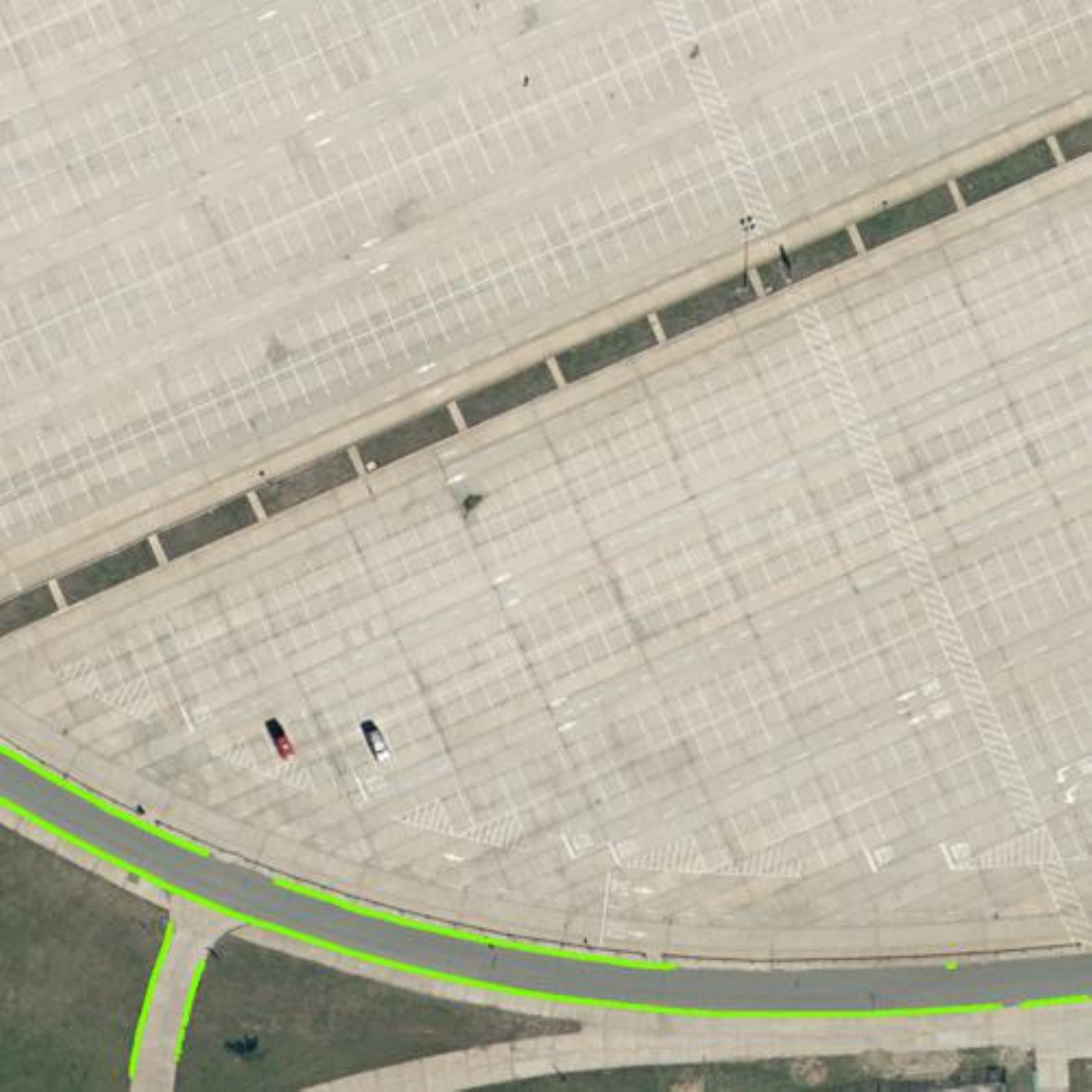}
        \end{subfigure}\vspace{.6ex}
        \begin{subfigure}[t]{\textwidth}
            \includegraphics[width=\textwidth]{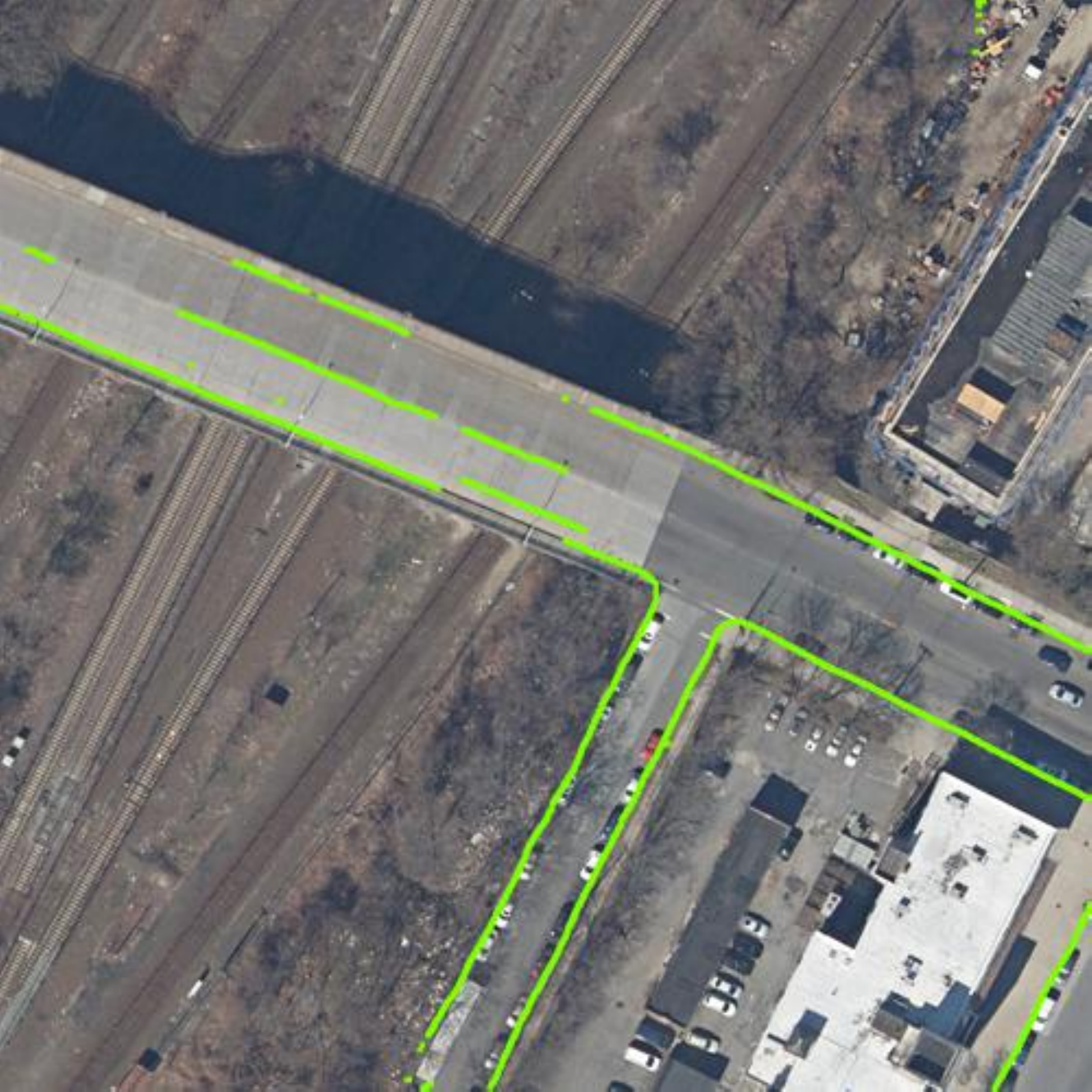}
        \end{subfigure}\vspace{.6ex}
        \begin{subfigure}[t]{\textwidth}
            \includegraphics[width=\textwidth]{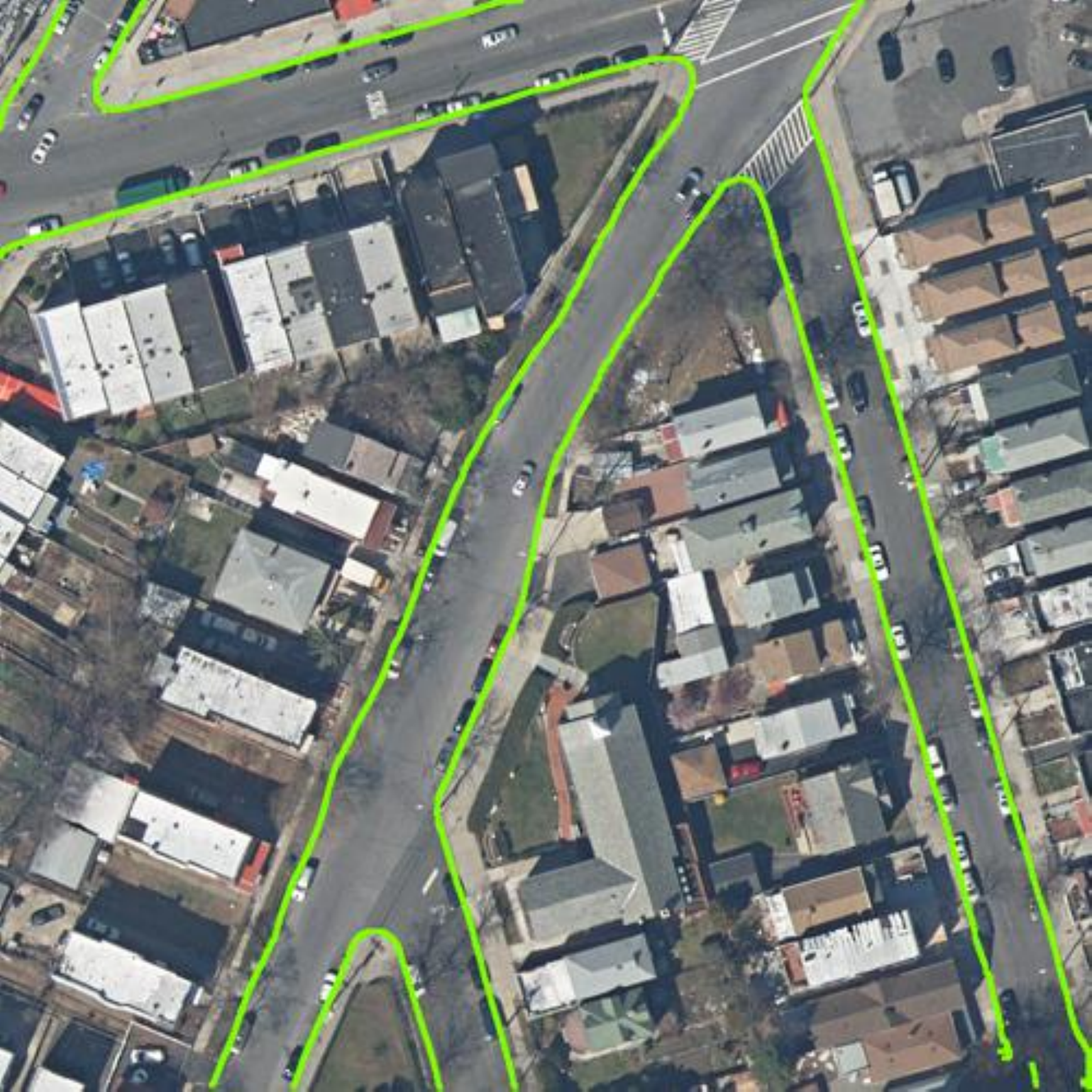}
        \end{subfigure}
        \caption{\footnotesize{Balance CE \cite{acuna2019devil}}}
    \end{subfigure}
    \hfill
    \begin{subfigure}[t]{0.1375\textwidth}
        \begin{subfigure}[t]{\textwidth}
            \includegraphics[width=\textwidth]{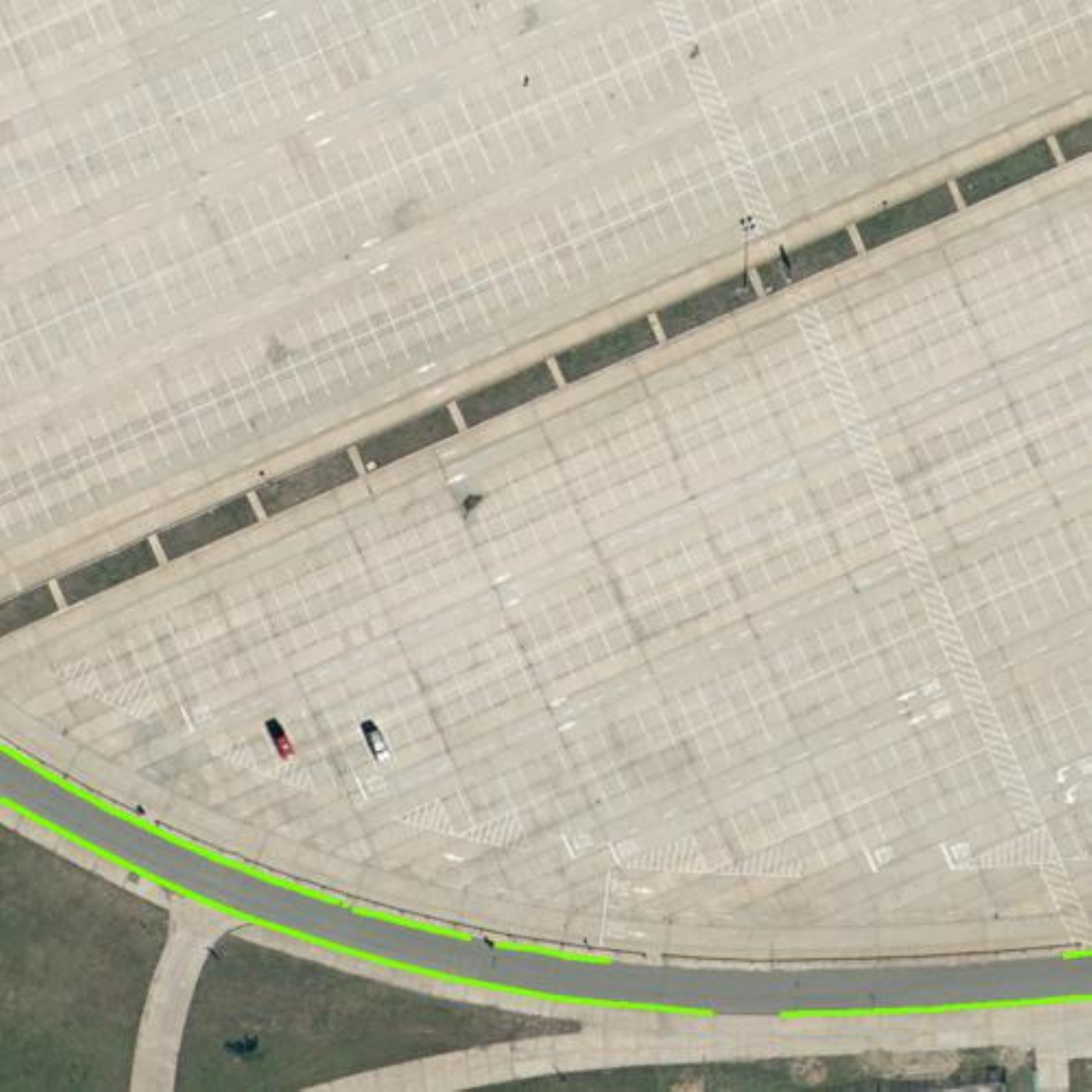}
        \end{subfigure}\vspace{.6ex}
        \begin{subfigure}[t]{\textwidth}
            \includegraphics[width=\textwidth]{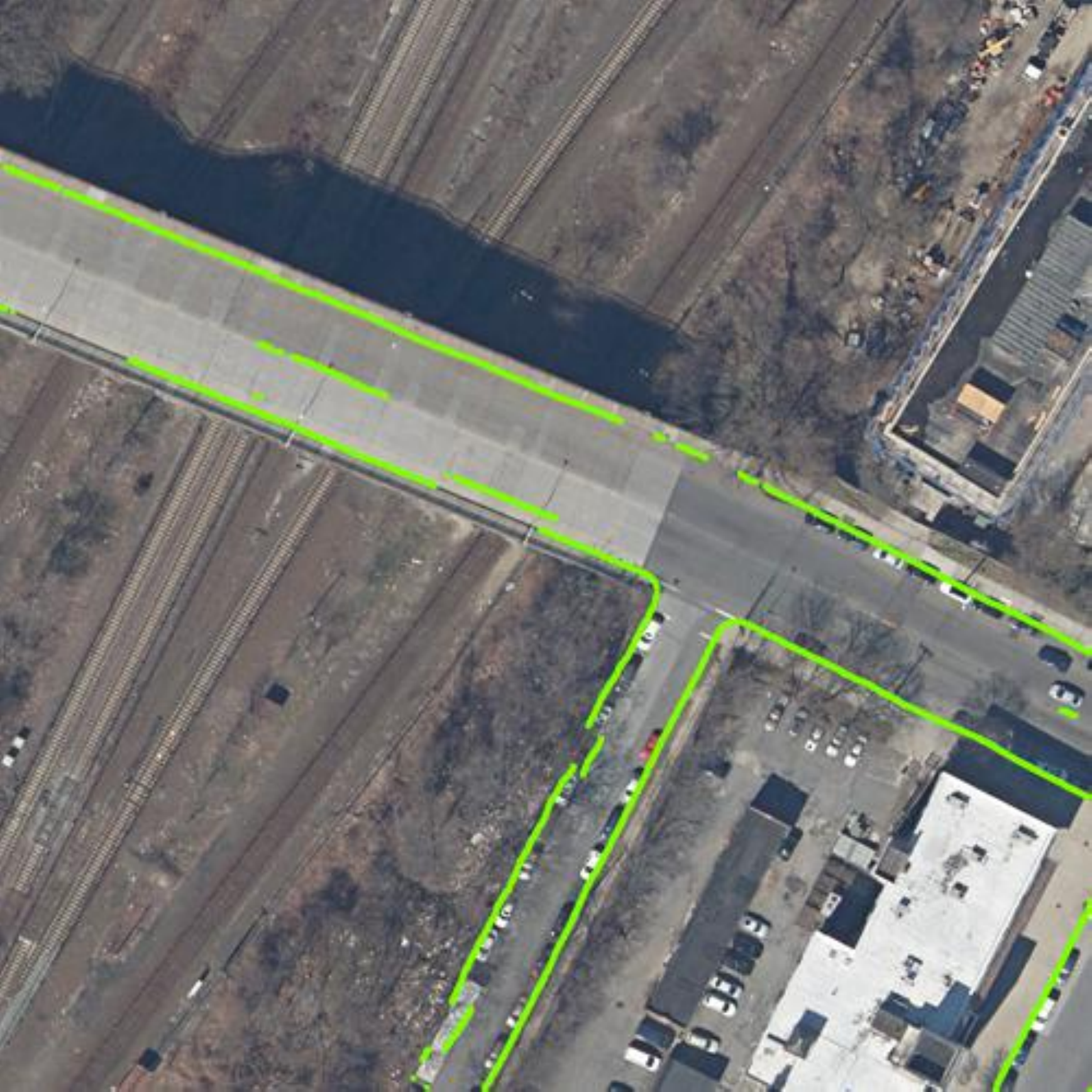}
        \end{subfigure}\vspace{.6ex}
        \begin{subfigure}[t]{\textwidth}
            \includegraphics[width=\textwidth]{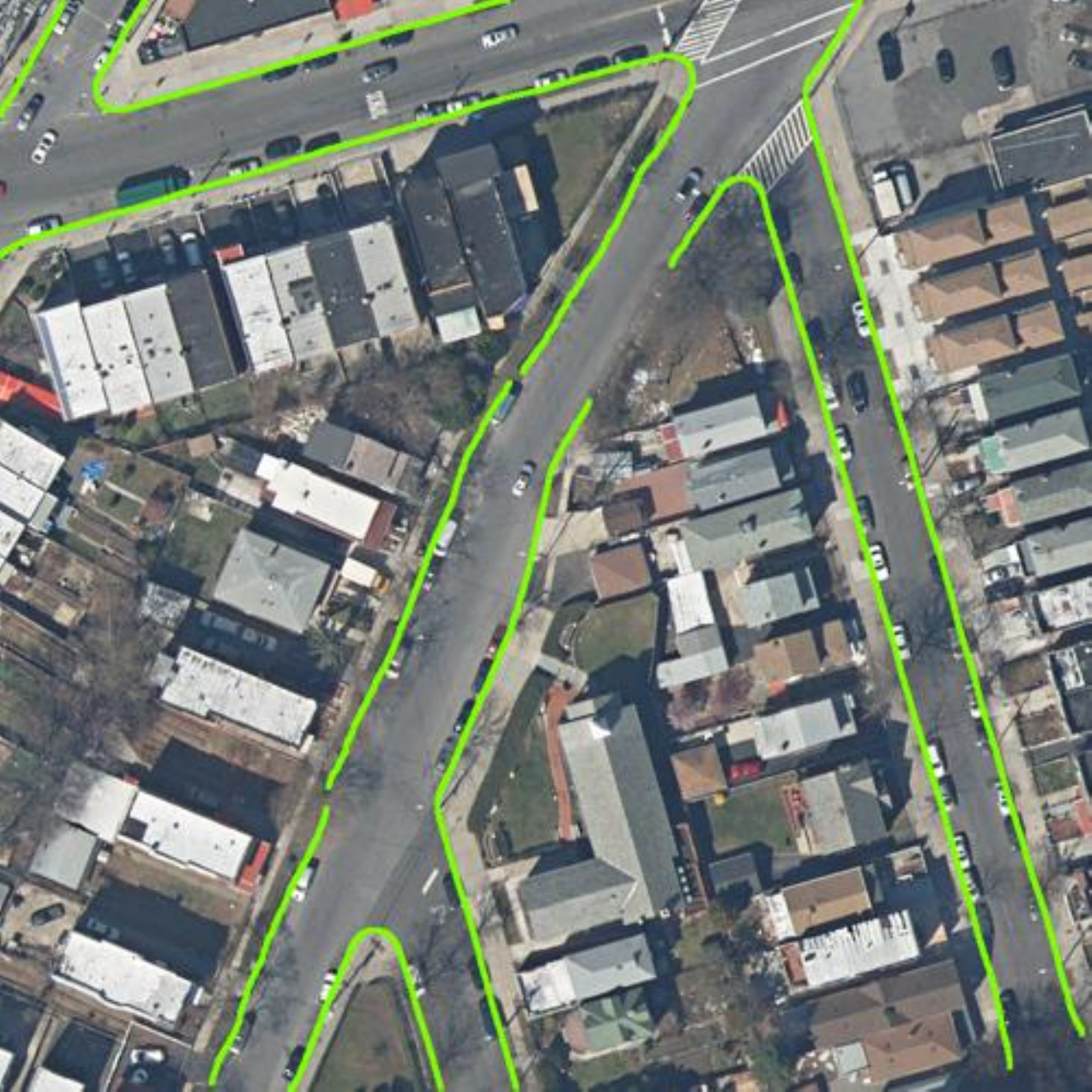}
        \end{subfigure}
        \caption{\footnotesize{Dice loss \cite{milletari2016vnet}}}
    \end{subfigure}
    \hfill
    \begin{subfigure}[t]{0.1375\textwidth}
        \begin{subfigure}[t]{\textwidth}
            \includegraphics[width=\textwidth]{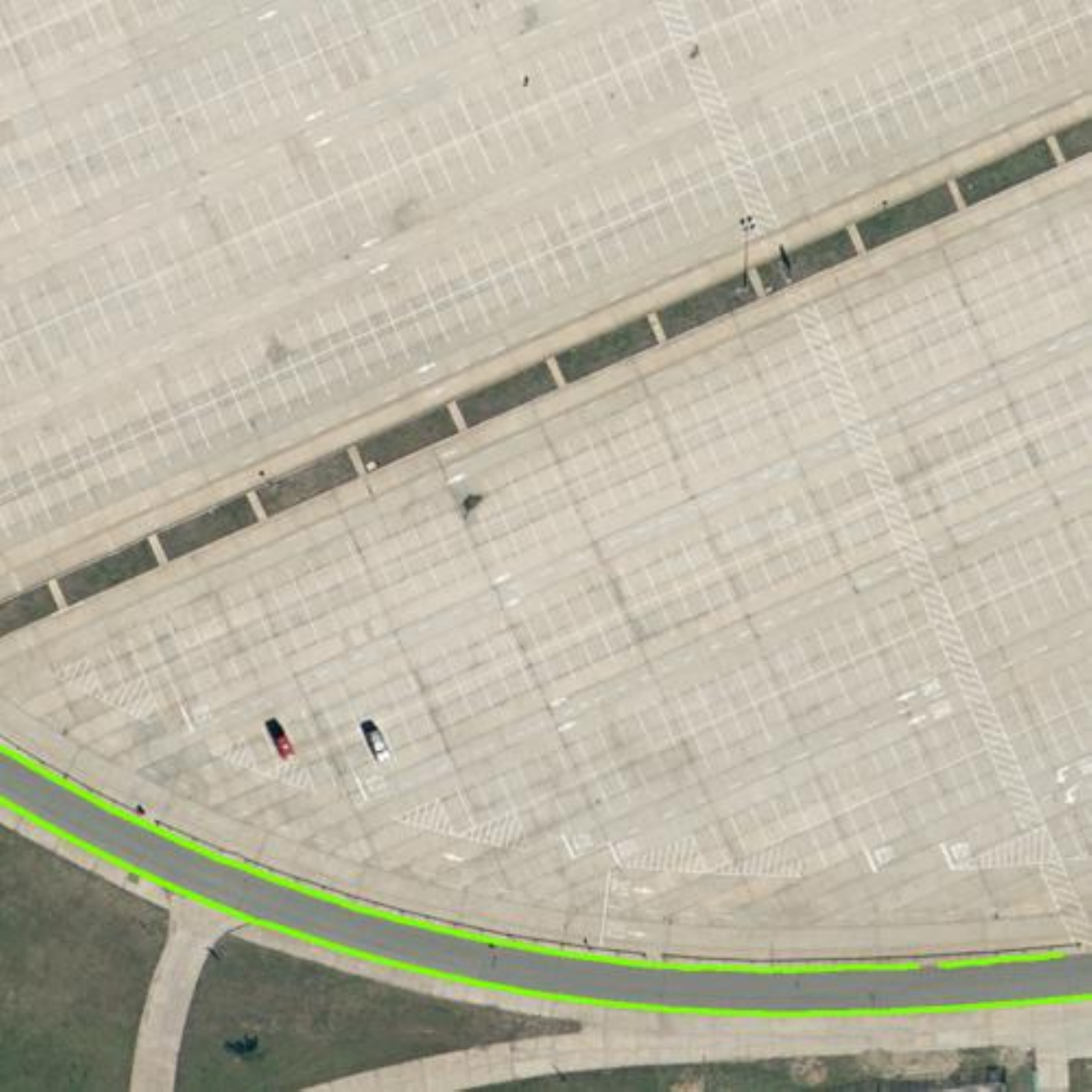}
        \end{subfigure}\vspace{.6ex}
        \begin{subfigure}[t]{\textwidth}
            \includegraphics[width=\textwidth]{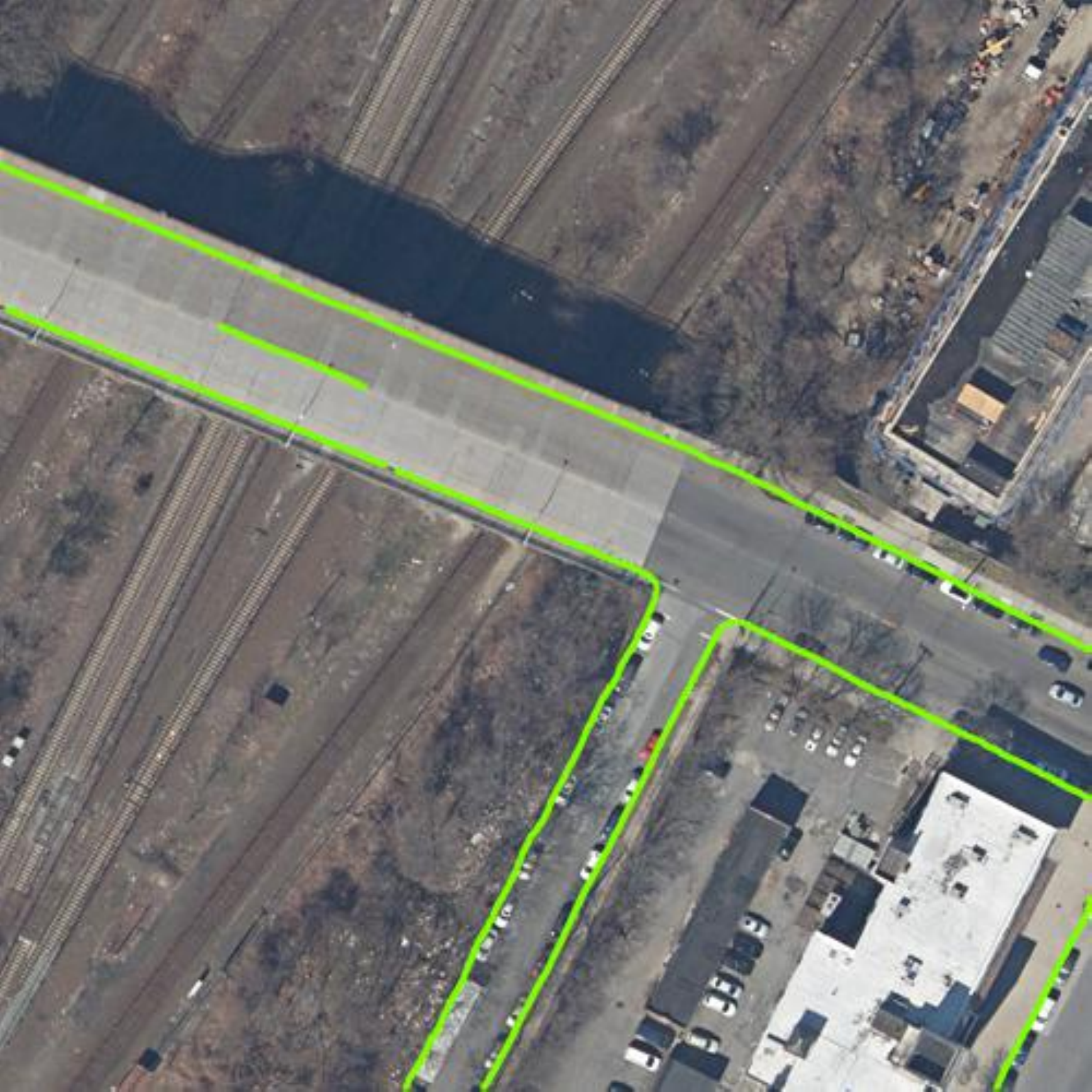}
        \end{subfigure}\vspace{.6ex}
        \begin{subfigure}[t]{\textwidth}
            \includegraphics[width=\textwidth]{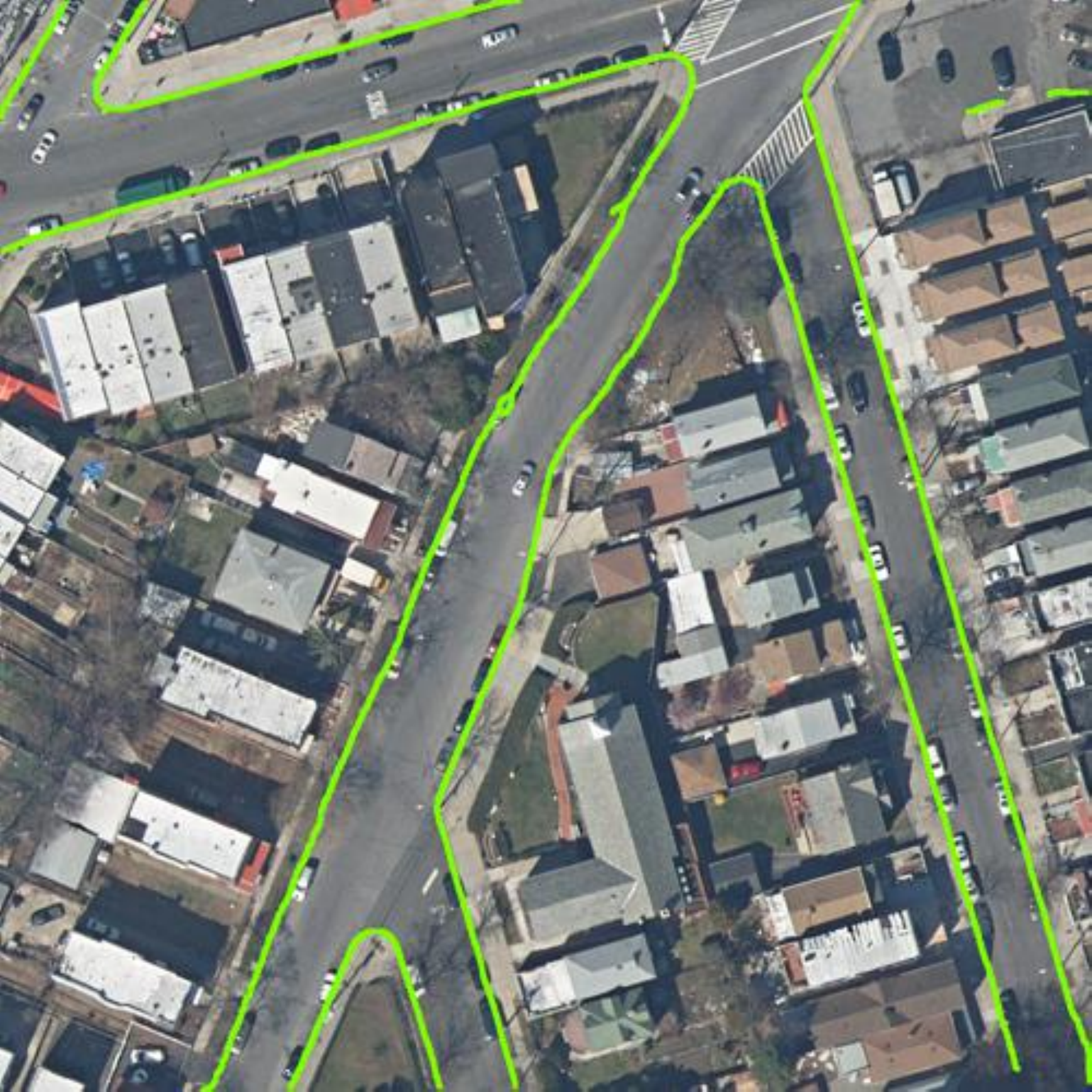}
        \end{subfigure}
        \caption{\footnotesize{Ours}}
    \end{subfigure}
    \caption{Sample demonstrations of the experimental results. (a) Ground-truth labels (cyan lines). (b)-(f) The final road curb skeletons of baselines. (g) The final road curb skeletons of CP-loss (green lines). Each row represents the experiment results of one testing image. From the visualization, we can see that the segmentation output of the network trained with CP-loss outperforms all the baselines, considering that it has better connectivity and fewer ghost skeletons. For better visualization, road curb lines in this figure are widened but they are actually of one-pixel-length. Please zoom in for details.}
    \label{fig_qualitative}
\end{figure*}
\subsection{CP-loss}\label{cploss}
\begin{algorithm}[t]
\KwInput{Segmentation result $S$, ground-truth label $GT$}
\Begin{
$S_{Bin}\gets (S>\tau_{Bin})$\\
$Skel_{P}\gets$ Skeletonization$(S_{Bin})$ \\
$Skel_{G}\gets$ Skeletonization$(GT)$ \\
$Skel_{fG}\gets Skel_{G}\cdot $\textit{far\_region}$(Skel_{P},\Delta)$\\
$Skel_{fP}\gets Skel_{P}\cdot $\textit{far\_region}$(Skel_{G},\Delta)$\\
\For{$x_i\in S$}{
            $u_i\gets [1 + exp[-\frac{\text{\textit{min\_dis}}(x_i,Skel_{fG})}{\sigma}]-p_i]^2$\\ 
            $v_i\gets [exp[-\frac{\text{\textit{min\_dis}}(x_i,Skel_{fG}+Skel_{fP})}{\sigma}]+p_i]^2$\\ 
            $\beta_i\gets \frac{1}{4}[1+exp[-\frac{\text{\textit{min\_dis}}(x_i,Skel_{fG}+Skel_{fP})}{\sigma}]-\frac{p_i}{2}]$\\
        }
$L_{CE}\gets \sum_i [-u_ig_ilog(p_i)-v_i(1-g_i)log(1-x_i)]$\\
$L_{Dice} = 1-2\frac{\sum_i \beta_ip_ig_i}{\sum_i (\beta_ip_i)^2+\sum_ig_i^2}$\\

\Return $L_{CE}+L_{Dice}$
}
\caption{CP-loss}
\label{alg1}
\end{algorithm}
Different from pixel-level errors, disconnectivity is hard to formulate merely based on discrete pixels. Thus CP-loss relies on morphological skeletons $Skel_{P}$ and $Skel_{G}$, which capture the structural information of the whole image. The algorithm to calculate CP-loss is shown in Alg. \ref{alg1}.

In CP-loss, the disconnectivity is measured by the failed-retrieved skeleton $Skel_{fG}$. The more pixels in $Skel_{fG}$, the worse connectivity is. However, $Skel_{fG}$ cannot be directly located based on raw $Skel_{P}$ and $Skel_{G}$, because skeletons are of one-pixel-width and the mismatch between $Skel_{P}$ and $Skel_{G}$ cannot be perfectly avoided. Therefore, we soften $Skel_{fG}$ and define a pixel $x_i$ in $Skel_{G}$ as failed-retrieved if its distance to $Skel_{P}$ is larger than a threshold $\Delta$. If $\Delta=1$, $Skel_{fG}$ is directly obtained from the raw $Skel_G$ and $Skel_P$. For computation convenience, we define a function \textit{far\_region}$(Skel',\delta)$ to find all the pixels whose shortest distance to the skeleton $Skel'$ is larger than $\delta$. In this way, we can obtain the soft $Skel_{fG}$, whose pixels are outside $\Delta$ distance to $Skel_{P}$. The soft $Skel_{fP}$ can be obtained similarly and it is used to punish ghost skeletons that represent false-positive predictions.

The original $L_{CE}$ is a pixel-level loss function and cannot effectively grasp structural information. To make it aware of disconnectivity, we add a weight coefficient $u_i$ or $v_i$ calculated from skeletons to each pixel $x_i$ in the segmentation result $S$, where $u_i$ is for foreground pixels ($g_i=1$) and $v_i$ is for background pixels ($g_i=0$). For foreground pixels, we need to concentrate on $Skel_{fG}$, so pixels closer to $Skel_{fG}$ should be assigned with larger $u_i$. The function to measure the distance is a Gaussian function $exp[-\frac{\text{\textit{min\_dis}}(x_i,Skel_{fG})}{\sigma}]$, where $\sigma$ is a hyper-paramter and function \textit{min\_dis$(x,\Omega)$} calculates the shortest distance between a pixel $x$ and a set $\Omega$:
\begin{equation}
    \text{\textit{min\_dis}}(x,\Omega) = min(\{\lVert x-x_i\rVert_2|\forall x_i \in \Omega\}).
\end{equation}
In this way, the closer the foreground pixels to $Skel_{fG}$ the larger weights they will receive. Since Focal loss has a tremendous ability to focus on harder samples, it is considered when designing $u_i$. Then, we combine the distance function and Focal loss by addition, which has better properties than multiplication (please refer to the supplementary file for more details). For background pixels, we want to not only emphasize connectivity but also avoid ghost skeletons. So the background pixel closer to either $Skel_{fG}$ or $Skel_{fP}$ receives larger weights. After calculating $u_i$ and $v_i$, a new weighted CE loss function aware of disconnectivity and ghost skeletons is obtained.

To further enhance connectivity and avoid ghosts, $L_{Dice}$ is also modified by assigning Euclidean distance-based weights $\beta_i$ for every pixel. Since $L_{Dice}$ processes the foreground and background at the same time, $\beta_i$ shares the same distance function with $v_i$ (i.e., both $Ske_{fG}$ and $Skel_{fP}$ are considered). Inspired by \cite{li2020dice}, the design of $\beta_i$ mimics Focal loss to focus on harder samples. Similar to weighted $L_{CE}$ designed above, the distance function is combined with Focal loss through addition. In Alg. \ref{alg1}, please note that $p_i$ in $\beta_i$ is divided by 2 and the whole $\beta_i$ is divided by 4. These two divisors are tuned to make the gradient of $L_{Dice}$ more appropriate for our task, otherwise more false-positive or false-negative predictions would occur. Please refer to the supplementary file for more details. 



Finally, the weighted $L_{CE}$ and weighted $L_{Dice}$ are summed up as the final CP-loss, which takes the advantages from both of them. 

\section{Experimental Results and Discussions}\label{experiment}
\subsection{Dataset}
Currently, there exists no available datasets or benchmarks about road curb detection yet. But in the New-York-City planimetrics (NYC-planimetrics) dataset \cite{nyc_dataset} provided by NYC OpenData, road curbs are annotated as a new feature in 2016. There are 2,049 high-resolution aerial images in this dataset, and each of them is $5000\times5000$-sized and has four channels, including red, green, blue and infrared. All the images have 0.5 ft/pixel$\approx$15.2 cm/pixel resolution. The road curb annotations are in the form of polylines and the vertices are recorded by the WGS84 coordinate system. Because the raw data cannot be directly used for our task, we create our own dataset from this public-available dataset.

We split each image into 25 $1000\times1000$ patches together with corresponding annotations, and convert the annotations from the WGS84 coordinate to the image coordinate. Patches that have no road curbs or have inappropriate annotations are removed. After the above pre-processing, we select around 18,000 image patches for our experiment. Among them, 17,000 images are used for training and the rest images are for testing.

 \begin{figure*}[t]
    \begin{subfigure}{.24\textwidth}
        \includegraphics[trim={0 0cm 1.5cm 2cm},clip,width=\linewidth]{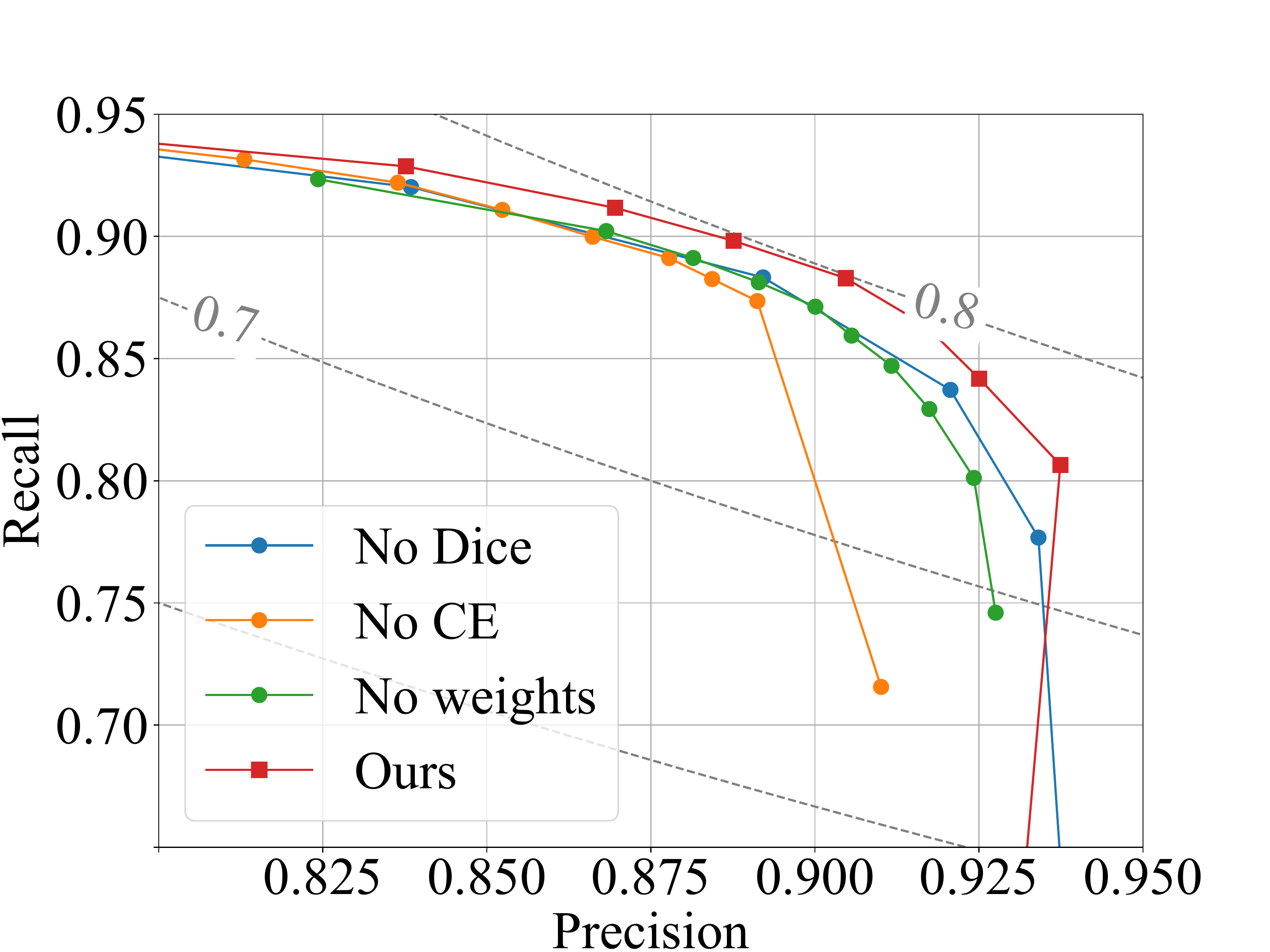}\caption{Ablation P-R curve}
    \end{subfigure}\hfill
    \begin{subfigure}{.24\textwidth}
        \includegraphics[trim={0 0cm 1.5cm 2cm},clip,width=\linewidth]{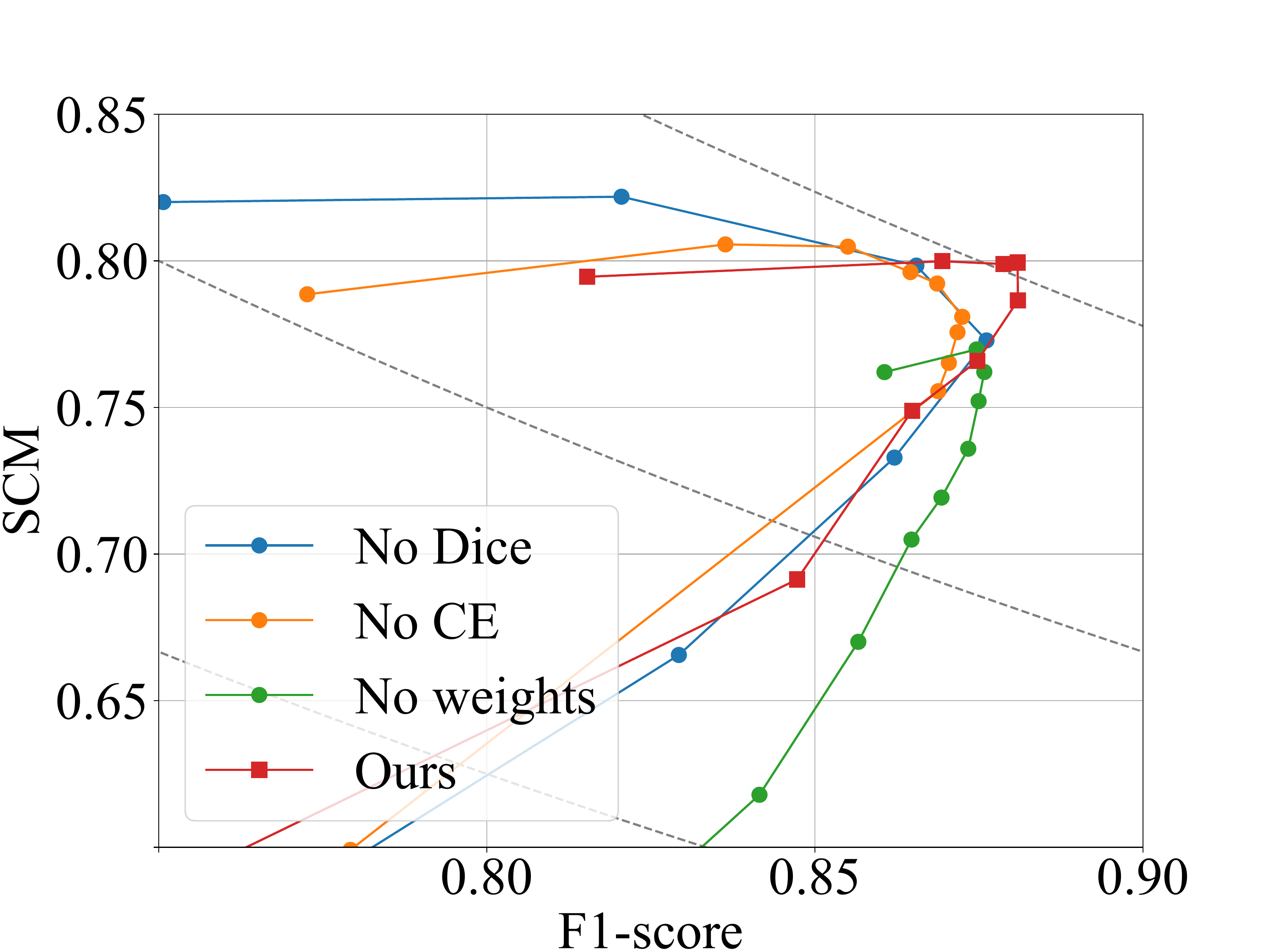}\caption{Ablation F-S curve}
    \end{subfigure}
    \begin{subfigure}{.24\textwidth}
        \includegraphics[trim={0 0cm 1.5cm 2cm},clip,width=\linewidth]{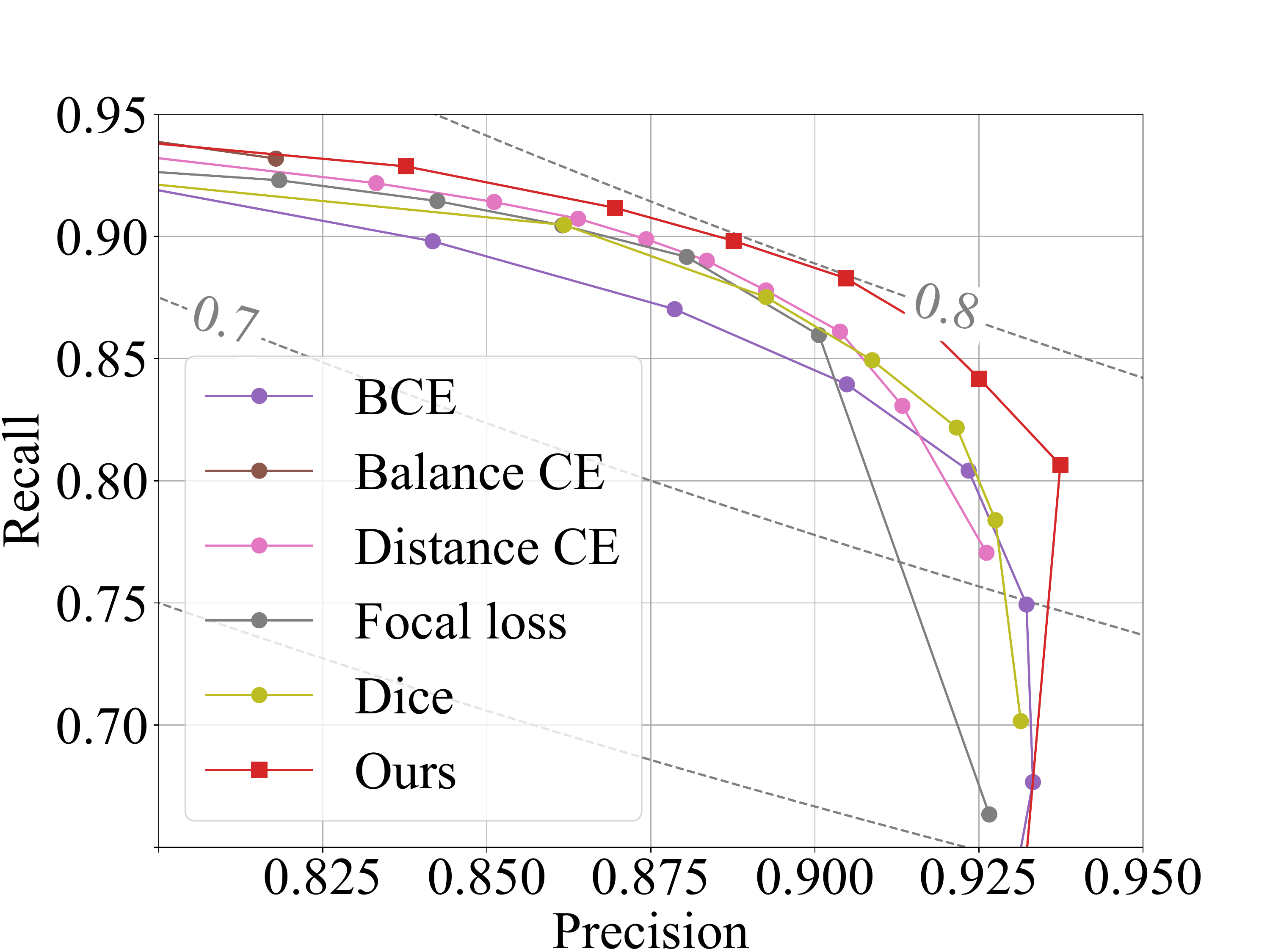}\caption{Comparative P-R curve}
    \end{subfigure}\hfill
    \begin{subfigure}{.24\textwidth}
        \includegraphics[trim={0 0cm 1.5cm 2cm},clip,width=\linewidth]{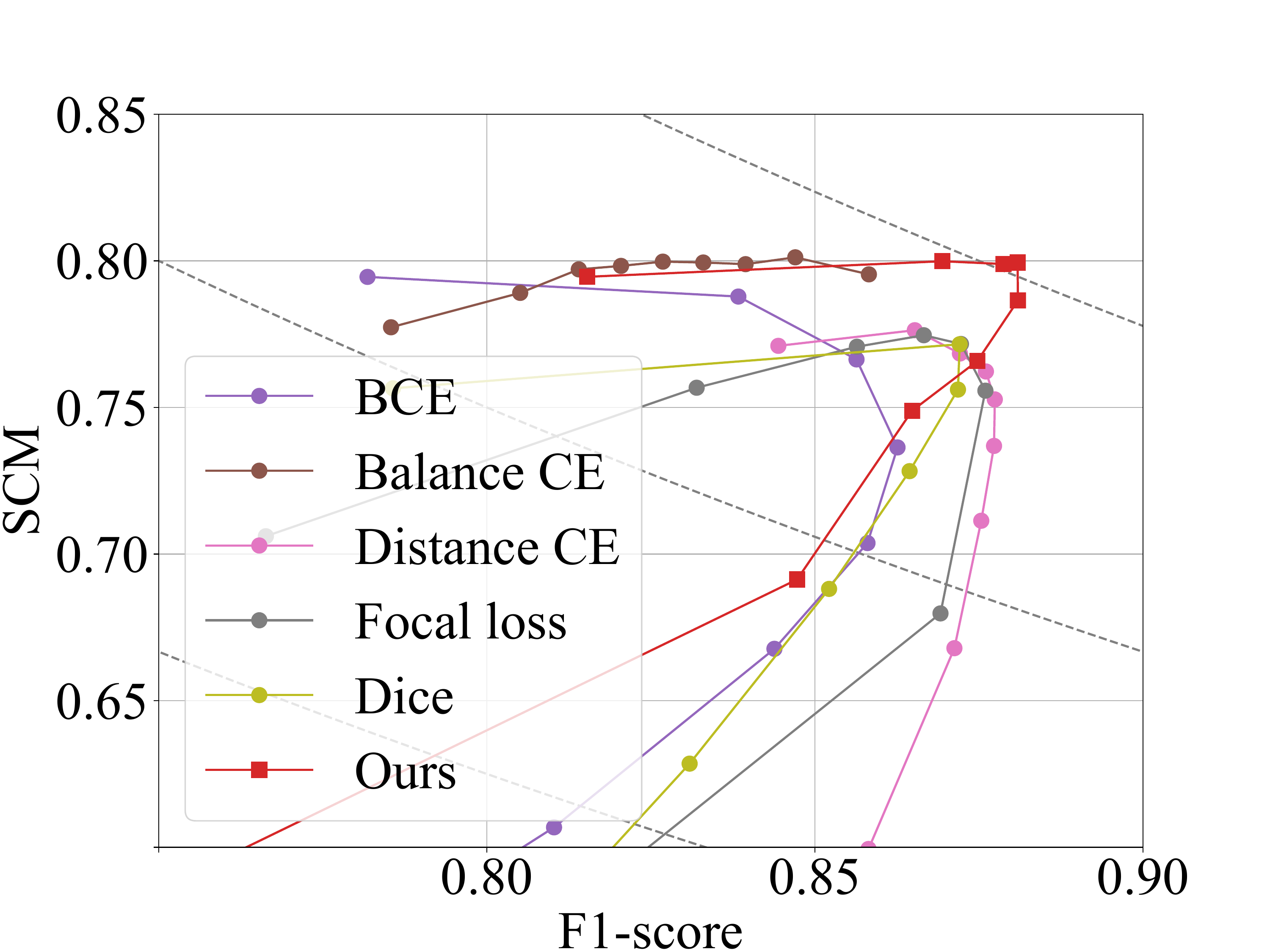}\caption{Comparative F-S curve }
    \end{subfigure}
    \caption{Statistical curves of the experiment results. These curves are obtained by binarizing the segmentation outputs with different threshold $\tau$. P-R curve illustrates the relationship between precision and recall as $\tau$ varies, and F-S curve is about F1-score and SCM. For both of these two categories of curves, being closer to the upper right corner indicates superiority. From the curves, we can see that CP-loss has better performance than any other baseline or variant. This figure is better viewed in color. Please zoom in for details.}
    \label{curves}
\end{figure*}

\subsection{Implementation process}
In our experiment, UNet serves as the segmentation network and we train it with different loss functions. For baseline loss functions, UNet is trained from randomly initialized parameters for 30 epochs. Due to the existence of the skeletonization operation in CP-loss, if we start the training from random initialization, the training process will become very slow and ineffective. Thus we train CP-loss and its variants from checkpoints pre-trained by binary CE. During testing, after the segmentation results are obtained, we binarize the results with a threshold $\tau$ and skeletonize them as the final output. For each loss function, the post-processing is done with multiple $\tau$ and then we obtain two statistical curves (i.e., precision-recall (P-R) curve and F1-score-SCM (F-S) curve) which are shown in Fig. \ref{curves}. The learning rate is $10^{-4}$ with a $10^{-5}$ decay rate. During the experiments, $\sigma$ is set to $100$ through tuning. The calculation of CP-loss could be well paralleled, thus in our experiment it is implemented by CUDA, and the efficiency is guaranteed.

\subsection{Evaluation metrics}
In past works about structural predictions, such as road-network extraction \cite{he2020sat2graph} and road-lane detection \cite{homayounfar2019dagmapper}, the obtained results are evaluated by pixel-level metrics including precision, recall and F1-score, as well as structure-aware metrics, such as APLS \cite{vanetten2019spacenet} and Connectivity \cite{homayounfar2019dagmapper}. 
In our experiments, these metrics are calculated based on binary skeletons instead of probabilistic maps. After obtaining wrong skeletons $Skel_{fG}$ and $Skel_{fP}$, we can have the correct skeletons $Skel_{tG}=Skel_G-Skel_{fG}$ and $Skel_{tP}=Skel_P-Skel_{fP}$. Then, the pixel-level metrics can be calculated by the following equations:
\begin{equation}
    \begin{aligned}
        P = \frac{|Skel_{tP}|}{|Skel_P|}, 
        R =\frac{|Skel_{tG}|}{|Skel_G|},
        F1 =\frac{2P\cdot R+\textit{s}}{P+R+\textit{s}}
    \end{aligned}
\end{equation}
where $P$ is precision, $R$ is recall and $\textit{s}$ is a small constant number to prevent zero denominators. $|\cdot|$ represents the number of pixels of a skeleton.

APLS is good for evaluating large shapes with complicated topologies, like road-network, but is not a proper choice for road curb detection, whose output is of simple shape and might not be connected. Inspired by Connectivity \cite{homayounfar2019dagmapper}, we propose a new metric named skeleton-connectivity-measure (SCM) to evaluate the connectivity of the obtained skeleton $Skel_{P}$. For each road curb instance $Skel_G^i$ in $Skel_{G}$, we find the corresponding true-positive predictions as $Skel_{tP}^i$, where $Skel_{tP}^i=\{x_i|{min\_dis}(x_i,Skel_{G}^i)<\Delta,\forall x_i \in Skel_P\}$. Similarly, the truly-retrieved $Skel_G^i$ denoted as $Skel_{tG}^i$ can be obtained. Ideally, for each instance, there should be one and only one predicted skeleton (i.e., $Skel_{tP}^i$ should be a single connected skeleton). But when the prediction has errors, there might be several separated skeleton segments in $Skel_{tP}^i$, and we record the number of separated segments of $Skel_{tP}^i$ as $n_i$. Then $\frac{1}{n_i}$ can be used to measure the connectivity of the predicted road curb instance. After getting $\frac{1}{n_i}$, we assign a weight $w_i=\frac{|Skel_{tG}^i|}{|Skel_G|}$ to each instance and sum them up as the final connectivity measurement. The calculation of SCM is shown in the following equation: 
\begin{equation}
    SCM=\sum_i \frac{|Skel_{tG}^i|}{|Skel_{G}|}\cdot c_i, c_i=\left\{
        \begin{array}{ll}
          \frac{1}{n_i}, &n_i\neq 0\\
          0 , &n_i = 0
        \end{array} \right.
\end{equation}

\subsection{Ablation study}
\begin{table}[t] 
\setlength{\abovecaptionskip}{0pt} 
\setlength{\belowcaptionskip}{0pt} 
\renewcommand\arraystretch{1.0} 
\renewcommand\tabcolsep{6.2pt} 
\centering 
\begin{threeparttable}
\caption{The quantitative results for the ablation study. The best results are highlighted in bold font. For all the metrics, larger values indicate better performance. We assess the weighted $L_{CE}$ (C), weighted $L_{Dice}$ (D), the weights of $L_{CE}$ ($W_{C}$) and the weight of $L_{Dice}$ ($W_{D}$).} 
\begin{tabular}{c c c c c c c c}
\toprule
C &  D & $W_{C}$ & $W_{D}$ & Precision & Recall & F1-score & $SCM$ \\ 
\midrule
&\checkmark&\checkmark&\checkmark &0.8661 & 0.8998 & 0.8725 & 0.7809  \\
\checkmark& &\checkmark&\checkmark &\textbf{0.8921} & 0.8832 & 0.8761 & 0.7729 \\
\checkmark&\checkmark&& & 0.8814 & 0.8912 & 0.8758 & 0.7621  \\
\midrule
\checkmark&\checkmark&\checkmark&\checkmark & 0.8696 &\textbf{0.9117}&\textbf{0.8809}&\textbf{0.7994} \\
\bottomrule 
\label{tab_ablation}
\end{tabular} 
\end{threeparttable}
\end{table}
In this section, the significance of the components of our loss function is evaluated. The quantitative results with different threshold $\tau$ are visualized in sub-figure (a) and (b) of Fig. \ref{curves}. The trade-off result is listed in Tab. \ref{tab_ablation}. 

Firstly, we build a variant loss function by removing the cross-entropy loss. Compared with cross-entropy loss, dice loss is less stable and is easy to be trapped in local optimal. From the experiment result, we notice an inferior performance of the network because of the lack of pixel-level supervision. Thus, the weighted cross-entropy loss is critical for satisfactory results.

Secondly, the dice loss is removed to make another variant. Then CP-loss becomes a weighted cross-entropy loss. Even though the weights of the cross-entropy loss enable the network to be aware of disconnectivity to some extent, it is still of pixel-level. Without the image-level dice loss, the segmentation result is affected. So the weighted dice loss is necessary for CP-loss.

Finally, we assess the importance of the weights used in CP-loss, including $u_i, v_i$ and $\beta_i$. 
Without these weights, CP-loss cannot obtain useful information on disconnections so that it fails to emphasize connectivity as expected. In Tab. \ref{tab_ablation}, we find that removing weights from CP-loss would seriously harm the final performance of the network. Therefore, the significance of our proposed disconnectivity-aware weights is confirmed.
  
\subsection{Comparative results}

\begin{table}[t] 
\setlength{\abovecaptionskip}{0pt} 
\setlength{\belowcaptionskip}{0pt} 
\renewcommand\arraystretch{1.0} 
\renewcommand\tabcolsep{8.5pt} 
\centering 
\begin{threeparttable}
\caption{The quantitative results for the comparative experiments. The best results are highlighted in bold font.} 
\begin{tabular}{c c c c c}
\toprule
Loss function&Precision&Recall&F1-score&$SCM$ \\ 
\midrule
BCE & \textbf{0.8786} & 0.8702 & 0.8626 & 0.7364  \\
Balance CE \cite{acuna2019devil} &0.8179 & \textbf{0.9318} & 0.8582 & 0.7954  \\
Distance CE \cite{caliva2019distance} & 0.8639 & 0.9073 & 0.8761 & 0.7623  \\
Focal loss \cite{lin2017focal} &0.8615 & 0.9045 & 0.8723 & 0.7717 \\
Dice loss \cite{milletari2016vnet} & 0.8618 & 0.9047 & 0.8721 & 0.7716  \\
\midrule
CP-loss  &0.8696 &0.9117&\textbf{0.8809}&\textbf{0.7994}\\
\bottomrule 
\label{tab_compare}
\end{tabular} 
\end{threeparttable}
\end{table}

To illustrate the superiority of our CP-loss, we compare CP-loss with multiple common-used loss functions for binary segmentation tasks. Similarly, the evaluation result with multiple $\tau$ is shown in sub-figure (c) and (d) in Fig. \ref{curves}. The trade-off result is listed in Tab. \ref{tab_compare}. Some example images are visualized in Fig. \ref{fig_qualitative}.

BCE refers to binary cross-entropy and it is simple but effective under most circumstances. However, BCE is a pixel-level loss function and cannot obtain structural information very well. Differently, dice loss is based on image-level calculation and can capture the image feature globally. But dice loss is less stable and is easy to be trapped in local optimal. Focal loss can enhance the results by assigning harder samples more weight, but it cannot explicitly emphasize disconnectivity, either. Therefore, the segmentation results merely based on aforementioned baselines are not sufficient for our task.

Balance CE is commonly used to handle the imbalance between the foreground and background \cite{acuna2019devil} by increasing the weight of foreground pixels. But for road curb segmentation whose shape is long and thin, it usually produces too thick predictions. As a result, the recall is greatly improved while the precision is seriously harmed. Moreover, it cannot specially focus on disconnectivities.

Distance CE means cross-entropy loss function with weights which are obtained by calculating the shortest distance between each pixel and the ground-truth label. Such a loss function is a good option for boundary segmentation tasks \cite{caliva2019distance}. But it concentrates on all the boundary pixels instead of those affected by disconnectivity. So distance CE does not achieve satisfactory performance on the road curb detection task.


\subsection{Failure cases and limitations}

 \begin{figure}[t]
        \begin{subfigure}[t]{0.155\textwidth}
            \includegraphics[width=\textwidth]{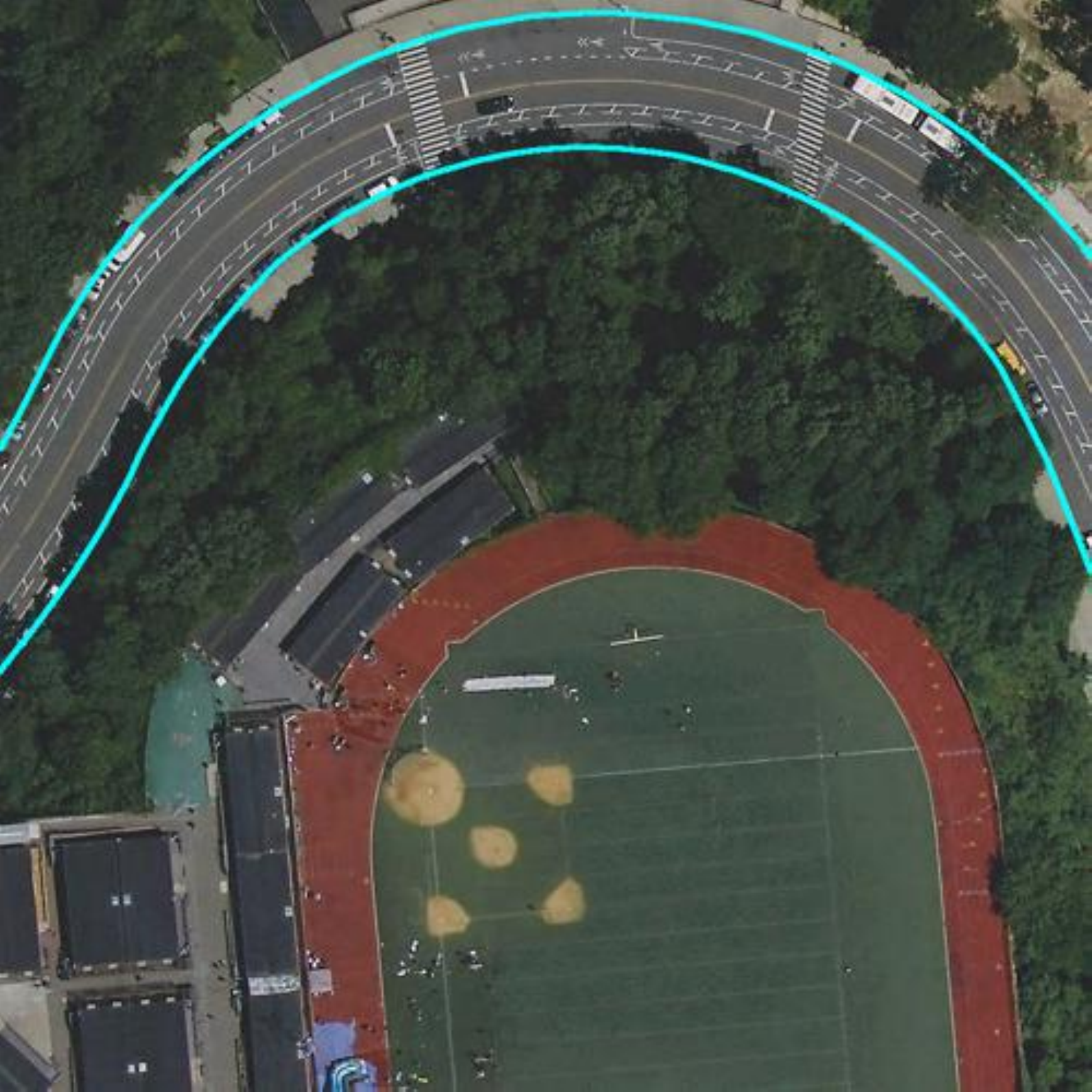}\caption{Ground-truth}
        \end{subfigure}
        \begin{subfigure}[t]{0.155\textwidth}
            \includegraphics[width=\textwidth]{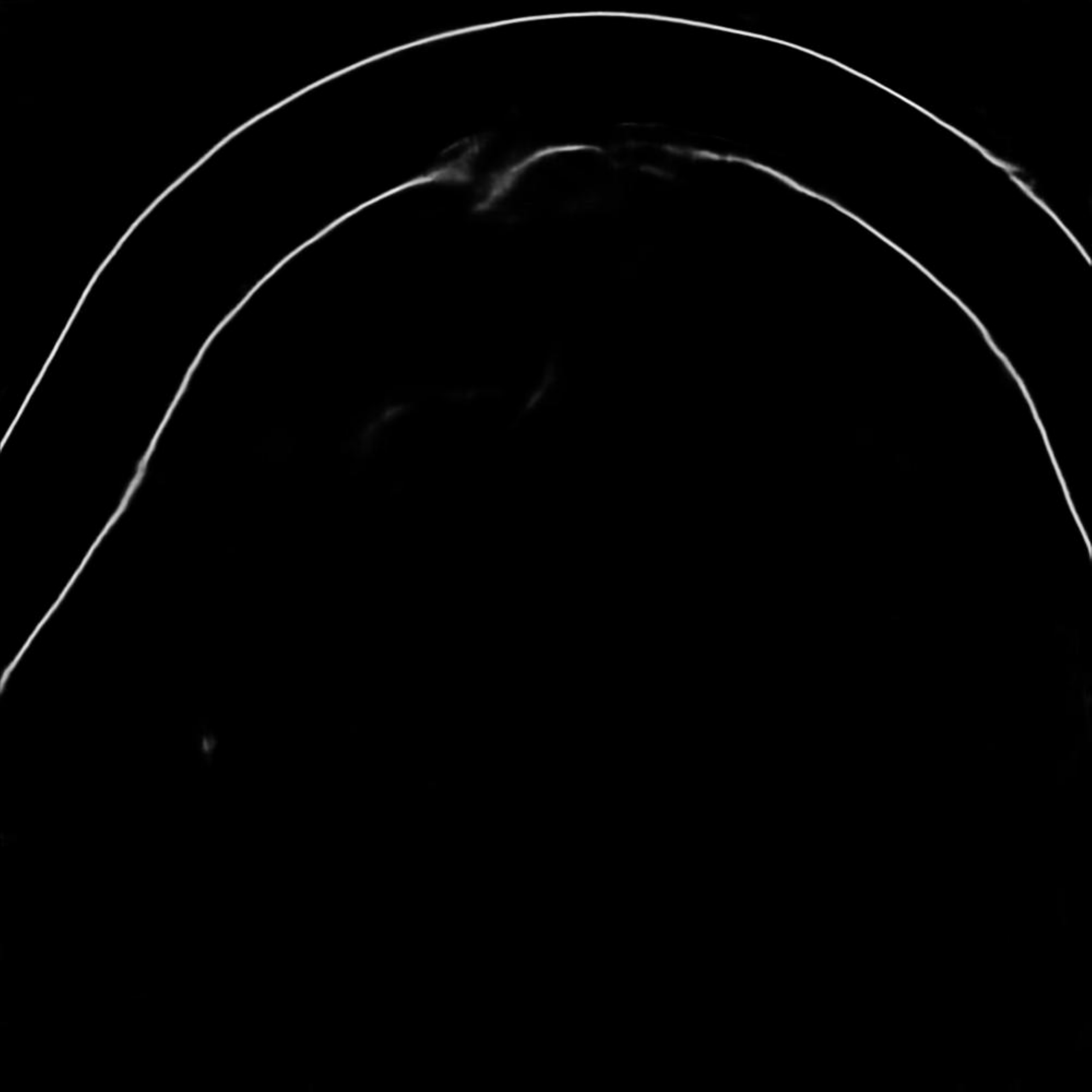}\caption{Segmentation}
        \end{subfigure}
        \begin{subfigure}[t]{0.155\textwidth}
            \includegraphics[width=\textwidth]{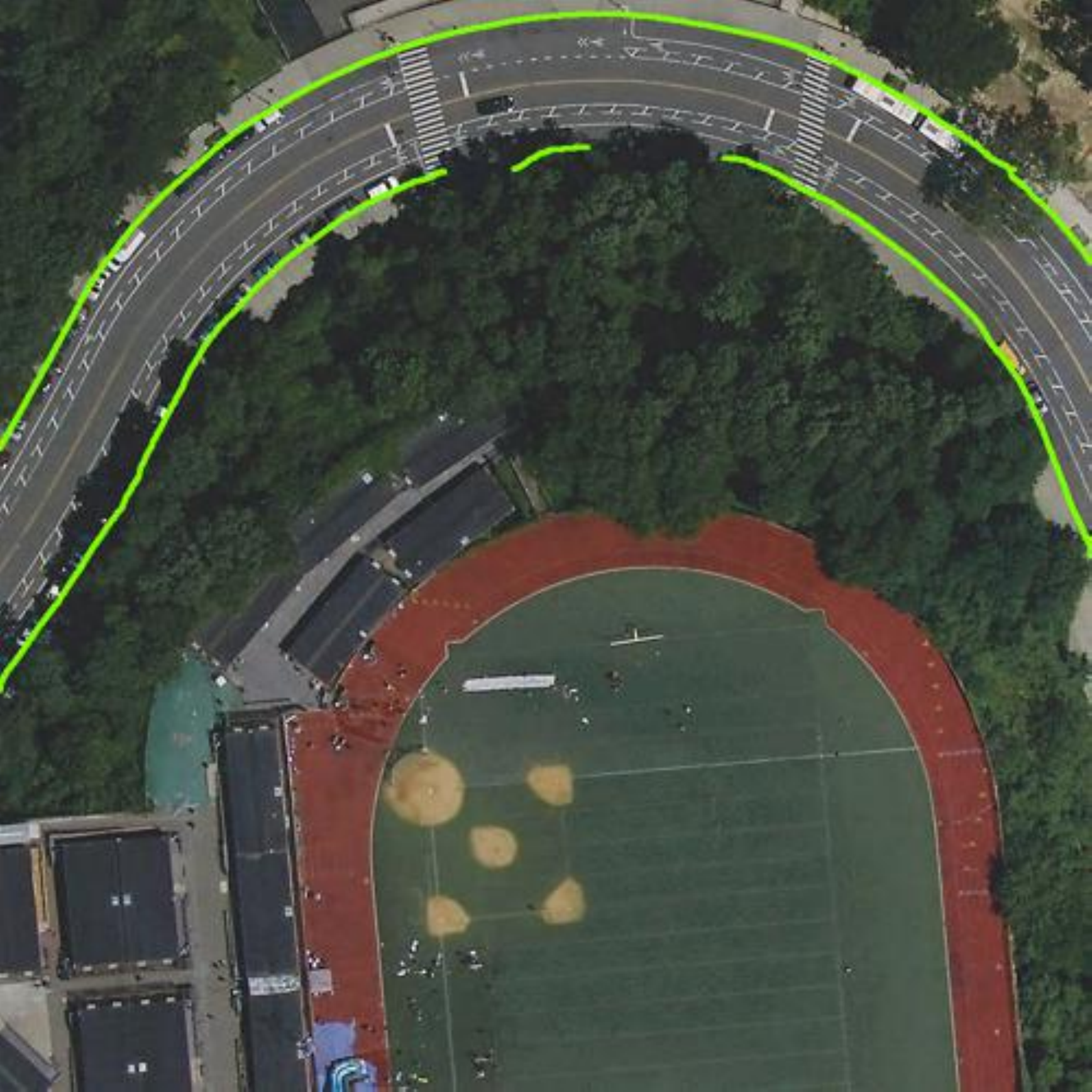}\caption{Skeleton}
        \end{subfigure}
        
    \caption{Qualitative demonstrations for a failure case of CP-loss.
    (a) The ground-truth label (cyan lines). (b) The segmentation probabilistic map. (c) The final obtained road curb skeleton (green lines).
    In this example, even though CP-loss outperforms other baselines, it cannot handle the disconnecivity issue very well since a large area of road curbs is seriously occluded by trees.
    (a) and (c) are widened for better visualization, but they are actually of one-pixel-width. }
    \label{fig_failure_case}
\end{figure}
Even though CP-loss can emphasize disconnectivity better than baseline loss functions, it cannot guarantee that outputs are free from disconnectivity issues, since deep learning models cannot cast hard constraints on the output. Therefore, there are still some failure cases caused by severe occlusion (e.g., occlusion by trees or shadows). An example is shown in Fig. \ref{fig_failure_case}. But compared with online detection, the occlusion issue is still greatly alleviated. Such a problem could be further relieved by designing better loss functions or using more powerful segmentation networks in the future.

\section{Conclusions and Future Works}
In this paper, we proposed an innovative loss function named CP-loss to handle the disconnectivity issue in road curb segmentation. CP-loss is a weighted combination of cross-entropy loss and dice loss. We first obtained the skeletons of both the predicted segmentation map and ground-truth label, then compared two skeletons and located areas where disconnectivity happens. Image pixels closer to these areas receive larger weights during training.
We created our own dataset by pre-processing a public dataset. To better evaluate the connectivity of the obtained road curbs, we designed a metric named SCM. The ablation studies and comparative experiments demonstrated the superiority of our CP-loss over commonly used loss functions for road curbs segmentation.
In the future, we plan to further refine our CP-loss to improve its effectiveness and efficiency. In addition, we will design new network structures incorporating our CP-loss for road curb detection.

\bibliographystyle{IEEEtran}
\bibliography{mybib}

\end{document}